\documentclass[10pt,twocolumn,letterpaper]{article}

\pdftrailerid{}
\usepackage[pagenumbers]{cvpr}

\usepackage{xspace}
\newcommand{\method}{SpecLook\xspace}
\newcommand{\methodfast}{\method-Fast\xspace}
\newcommand{\scout}{Scout\xspace}
\newcommand{\overseer}{Overseer\xspace}

\usepackage{graphicx}
\usepackage{colortbl}
\usepackage{multirow}
\usepackage{placeins}
\usepackage[skins]{tcolorbox}
\definecolor{promptbar}{RGB}{64,64,64}
\definecolor{promptbg}{RGB}{242,242,242}
\newtcolorbox{promptbox}[2]{enhanced, colback=promptbg, colframe=promptbar, coltitle=white,
  fonttitle=\small, title={#1\hfill #2}, arc=1.5mm, boxrule=0.5pt,
  left=4pt, right=4pt, top=2pt, bottom=2pt, toptitle=1pt, bottomtitle=1pt,
  fontupper=\footnotesize\raggedright, before skip=0pt, after skip=4pt}
\definecolor{oursrow}{RGB}{236,232,250}
\definecolor{grouprow}{RGB}{222,222,228}
\definecolor{gaingreen}{RGB}{27,122,61}
\definecolor{spdgreen}{RGB}{120,196,128}

\newcommand{\second}[1]{\underline{#1}}
\newcommand{\spd}[2]{\cellcolor{spdgreen!#1}#2}
\newcommand{\softrule}{\arrayrulecolor{black!30}\midrule[0.3pt]\arrayrulecolor{black}}
\usepackage{amsthm}
\newtheorem{proposition}{Proposition}

\definecolor{cvprblue}{rgb}{0.21,0.49,0.74}
\usepackage[pagebackref,breaklinks,colorlinks,allcolors=cvprblue]{hyperref}

\def\confName{CVPR}
\def\confYear{2027}

\title{\method: Speculating Where to Look for Fast and Reliable Agentic MLLMs}

\author{Xi Ding \qquad Jiawei Zhang\\
\raisebox{-0.3\height}{\includegraphics[height=1.3em]{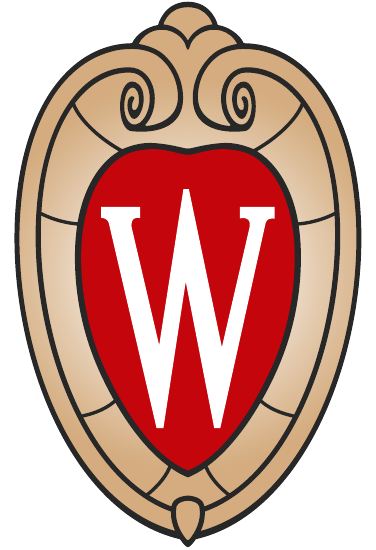}}\hspace{0.4em}University of Wisconsin--Madison}

\begin{document}
\maketitle

\begin{abstract}
Agentic multimodal large language models (MLLMs) that think with images interleave reasoning with zoom-in tool calls.
However, this sequential loop is slow, and its answers carry no reliability score.
Prior speculation of tokens, reasoning steps, or answers keeps the loop, and answer speculation also accepts confident errors.
We find that the loop mostly adds one zoom-in step, with which a small model is ahead of the agent on most splits.
Our key insight is to speculate the agent's next observation, which, unlike a drafted answer, can be verified, corrected, and used to test the answer.
We propose \textbf{\method}, a training-free speculate-and-verify framework with three levels.
A small model drafts a glimpse and re-answers from it, and the question ends if the answer is unchanged and confident.
Otherwise, the agent's large model verifies and corrects the glimpse without tools, and their answers are fused.
Only questions that still lack evidence reach the agent, in one forward pass that \methodfast skips, so no question runs the loop.
The trace scores every answer's reliability without labels.
Extensive experiments on V*, HR-Bench, and POPE show that both are more accurate than every baseline on every split.
\method reaches 87.2\% and 86.2\% mean accuracy with 7.7$\times$ and 4.7$\times$ speedups over the DeepEyes-7B and Thyme-RL agents, and \methodfast reaches 9.9$\times$ and 6.3$\times$.
\end{abstract}

\section{Introduction}
\label{sec:intro}

\begin{figure}[t]
  \centering
  \includegraphics[width=\linewidth]{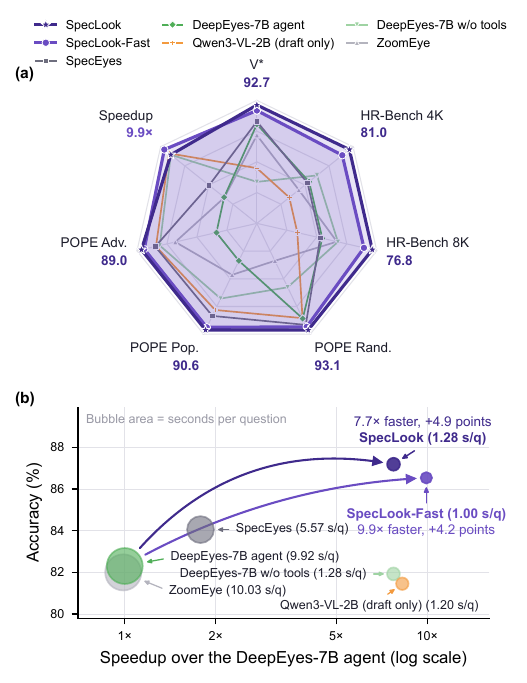}
  \caption{Accuracy and speed with DeepEyes-7B as the expert. (a) Accuracy on each split and speedup over the agent. (b) Mean accuracy over the six splits versus speedup.}
  \label{fig:teaser}
\end{figure}

\begin{figure}[t]
  \centering
  \includegraphics[width=\linewidth]{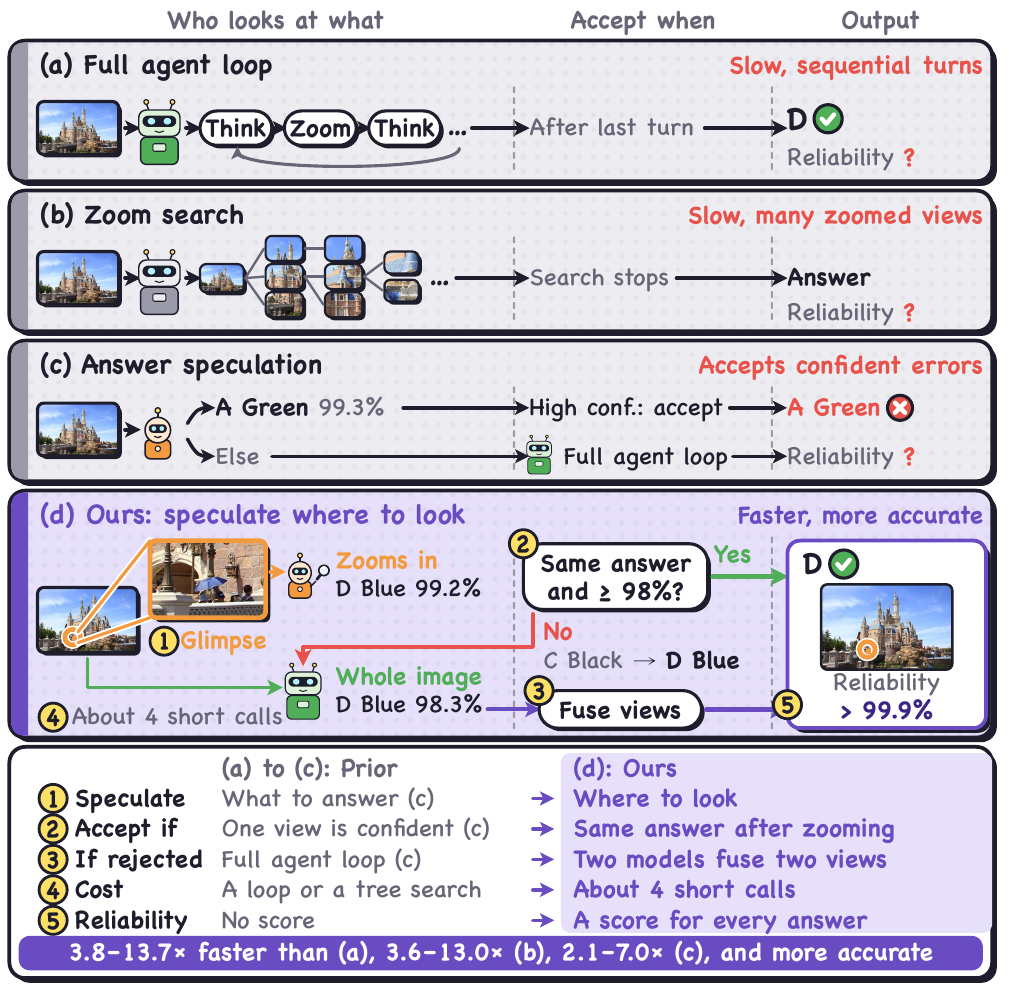}
  \caption{Three prior paradigms and \method on an HR-Bench 8K question. (a) DeepEyes-7B~\cite{zheng2026deepeyes}, (b) ZoomEye~\cite{shen2025zoomeye} (schematic), and (c) SpecEyes~\cite{huang2026speceyes}. Markers 1 to 5 match the bottom rows. Speedups use DeepEyes-7B as the expert.}
  \label{fig:comparison}
\end{figure}

\begin{figure}[t]
  \centering
  \includegraphics[width=\linewidth]{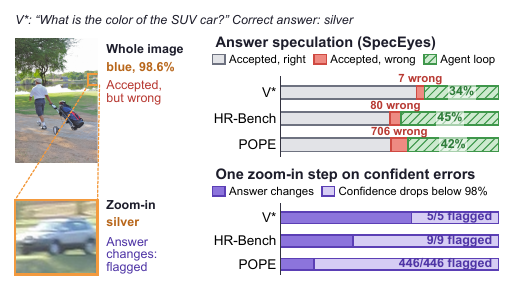}
  \caption{Top: Answer speculation by SpecEyes with DeepEyes-7B. Bottom: One zoom-in step by the same 2B model on its wrong drafts with a glimpse and at least 98\% confidence.}
  \label{fig:answer_speculation}
\end{figure}

Multimodal large language models (MLLMs) are moving from answering in a single pass to agentic inference with visual tools~\cite{yao2025agenticsurvey}.
Models that think with images, such as DeepEyes~\cite{zheng2026deepeyes} and Thyme~\cite{zhang2026thyme}, alternate between reasoning and a tool call that crops a region, and read the crop as a new observation~\cite{yao2023react,schick2023toolformer}.
The crop enlarges details that are too small in the whole image, which helps on questions about a small region~\cite{wu2024vstar,wang2025divide,li2024monkey,shi2024s2}.

This gain costs time and leaves a gap in reliability.
Every turn decodes long reasoning text and encodes a new crop, and turns run in sequence, so the cost of a question grows with its number of turns (\cref{fig:teaser}).
The final answer also has no reliability score, so its wrong answers go undetected.
Both problems call for a cheap test of whether an answer is right, which would let the agent skip its loop on the answers that pass and tell a user which answers to trust.

Speculation provides such a test. In speculative decoding, a cheap draft model proposes tokens and the large model verifies them~\cite{stern2018blockwise,leviathan2023fast,chen2023accelerating,cai2024medusa,li2024eagle}.
Speculation has since moved to reasoning steps~\cite{pan2025specreason} and to the final answer, but token and step speculation still run every turn of the loop.
SpecEyes~\cite{huang2026speceyes}, like a model cascade~\cite{viola2001rapid,chen2024frugalgpt}, accepts the confident answer of a small model on the questions that the large model marks as tool-free, and runs the agent loop on the rest (\cref{fig:comparison}).
Its only test is single-view confidence, so confident errors pass, and with DeepEyes-7B the full loop still runs on 34\% to 45\% of the questions (\cref{fig:answer_speculation}, top).

\textbf{What does the agent loop add to a single pass?}
We find that it mostly adds one zoom-in step, as a question in the released DeepEyes-7B runs typically makes one tool call.
When a small draft model, Qwen3-VL-2B~\cite{bai2025qwen3vl}, takes this step and reads its own glimpse, it is ahead of the agent on most splits and close to it on the others (\cref{sec:analysis}).
The step also supplies the test that a single view lacks, as every confident wrong draft with a glimpse changes its answer or is no longer confident after zooming (\cref{fig:answer_speculation}, bottom).
Answer speculation leaves tool calls by the draft model to future work~\cite{huang2026speceyes}, and zoom search~\cite{shen2025zoomeye} explores many views.
Our key insight is to move speculation one step earlier, from the agent's final answer to its next observation.
With or without a zoom by the draft model, a rejected answer sends the question to the loop, whereas a rejected region is replaced by the large model in one call, and the answer read from the region tests the whole-image answer.

We propose \method, a training-free speculate-and-verify framework in which each question stops at the first of three levels that settles it.
At the first level, a small MLLM, the \scout, answers from the whole image, drafts a glimpse, and re-answers from it, and the question ends when the answer is confident and unchanged after zooming, a certified exit, or confident with no glimpse drafted.
At the second level, the agent's large model, the \overseer, joins without its tools. As in speculative decoding, it verifies the glimpse and drafts its own on rejection, and a product of experts~\cite{hinton2002training} combines the \scout on the glimpse with the \overseer on the whole image, as a shared view correlates their errors.

At the third level, on questions with weak evidence, \method searches image tiles and, on multiple-choice questions, the agent answers in one forward pass, with the found regions in place of its tool calls.
This pass, on a small share of the questions, is the only agent call, so no question runs the loop, and \methodfast stops at the second level.
One drafted glimpse gives the exit test, the second view, and a trace that ranks every answer by its evidence.

Our contributions are as follows.
\begin{itemize}
  \setlength{\itemsep}{0pt}
  \item \textbf{A diagnosis of answer speculation.} We show that one zoom-in step, left to the agent loop by answer speculation, carries most of the agent's gain, can be drafted by a small model, and tests confident drafts on new evidence.
  \item \textbf{Observation-level speculation.} We propose \method, a training-free framework that speculates the agent's next observation instead of its final answer, so no question runs the loop. It fuses two models on different views, and we prove when a shared view lowers their oracle accuracy.
  \item \textbf{Label-free reliability.} We turn the speculation trace into an evidence region and a reliability score for every answer, with no extra model call. Every certified exit is correct, and the score detects wrong answers better than the acceptance score of SpecEyes on every benchmark.
  \item \textbf{Consistent gains in speed and accuracy.} On V*, HR-Bench, and POPE with DeepEyes-7B and Thyme-RL, \method is the most accurate on every split at 7.7$\times$ and 4.7$\times$ their speed, and \methodfast runs at 9.9$\times$ and 6.3$\times$ with higher accuracy than every baseline.
\end{itemize}

\section{Related Work}
\label{sec:related}

\textbf{High-resolution perception and zooming.}
MLLMs~\cite{radford2021clip,alayrac2022flamingo,li2023blip2,liu2023llava,liu2024improved,chen2024internvl,bai2025qwen25vl,bai2025qwen3vl} see an image through a bounded number of visual tokens, so a small object takes only a few.
High-resolution MLLMs encode slices or several scales~\cite{li2024monkey,guo2024llavauhd,shi2024s2}, and token pruning~\cite{chen2024fastv,zhang2025sparsevlm} or demo-free task vectors~\cite{ding2026two} cut visual tokens.
Glimpse-based models attend to regions in sequence~\cite{mnih2014recurrent,fu2017look} and choose where to zoom to save computation~\cite{gao2018dynamic,uzkent2020learning,wang2020glance}.
For MLLMs, training-free methods crop the most attended region~\cite{zhang2025mllms}, search a tree of regions~\cite{shen2025zoomeye}, merge patch descriptions~\cite{wang2025divide}, or retrieve crops~\cite{wang2025rap}.
These methods use the crop only to answer, while \method treats it as a draft observation that a second model verifies and that tests the draft answer.

\textbf{Agentic MLLMs that think with images.}
Visual programs compose vision modules~\cite{gupta2023visprog,suris2023vipergpt}, Visual Sketchpad~\cite{hu2024sketchpad} draws on the image, SEAL~\cite{wu2024vstar} adds a guided visual search model, and Visual CoT~\cite{shao2024visualcot} answers from the crop of a predicted key region.
With reinforcement learning~\cite{guo2025deepseekr1}, DeepEyes~\cite{zheng2026deepeyes} learns to call a zoom-in tool, Thyme~\cite{zhang2026thyme} generates and executes image-processing code, and Chain-of-Focus~\cite{zhang2025chainfocus} and Mini-o3~\cite{lai2026minio3} train adaptive and multi-turn visual search.
In all of them, one large model generates every step, while \method splits the zoom step between two models with no training and calls the agent in one forward pass, only on flagged questions.

\textbf{Speculative acceleration and cascades.}
Blockwise and speculative decoding~\cite{stern2018blockwise,leviathan2023fast,chen2023accelerating} draft tokens with a cheap model and verify them in one target-model pass, and later methods improve the drafter or the verification~\cite{zhou2024distillspec,sun2023spectr,cai2024medusa,li2024eagle,li2024eagle2,zhang2024draft,elhoushi2024layerskip,kim2023speculative,gagrani2024speculative,xia2024unlocking}.
Process verifiers score reasoning steps~\cite{lightman2024verify}, SpecReason~\cite{pan2025specreason} lets a small model draft steps that the large model scores, and Speculative Thinking~\cite{yang2025speculative} lets a large model guide a small one.
Cascades, early exits, and routers escalate on a low score~\cite{viola2001rapid,huang2018msdnet,xin2020deebert,schuster2022calm,chen2024frugalgpt,wang2022wisdom,gupta2024cascades,ding2024hybrid,aggarwal2024automix,ong2025routellm}, on inconsistent sampled answers~\cite{yue2024cascades}, or on an answer that changes after token pruning~\cite{zhao2025stitch}, all tests on the same or less visual evidence.
SpecEyes~\cite{huang2026speceyes} speculates the agent's answer in this way and runs the agent loop on every rejected question.
\method speculates the agent's next observation, exits only when the answer is unchanged under added evidence, a test none of these can make, and corrects a rejected glimpse in one call, with no loop.

\textbf{Fusion and reliability.}
A product of experts multiplies model distributions~\cite{hinton2002training,kittler1998combining}, and co-training~\cite{blum1998combining} assumes views that are conditionally independent given the label.
Deep ensembles estimate uncertainty~\cite{lakshminarayanan2017simple}, and modern networks are often miscalibrated~\cite{guo2017calibration}.
Selective prediction abstains on low confidence~\cite{hendrycks2017baseline,geifman2017selective,geifman2019selectivenet}, also for VQA~\cite{whitehead2022reliable} and with a selector trained on localized evidence~\cite{rodriguez2026sieves}, and the agreement of sampled answers tracks correctness~\cite{wang2023selfconsistency,kuhn2023semantic,manakul2023selfcheckgpt}.
Ensemble diversity is studied for differently trained classifiers~\cite{kuncheva2003measures}, while \method sets it at inference by the view assignment of two fixed models and obtains its exit test, fused views, and reliability score from one trace, with no training, no extra call, and no labels.

\section{Method}
\label{sec:method}

\begin{figure*}[t]
  \centering
  \includegraphics[width=\linewidth,height=0.44536\linewidth]{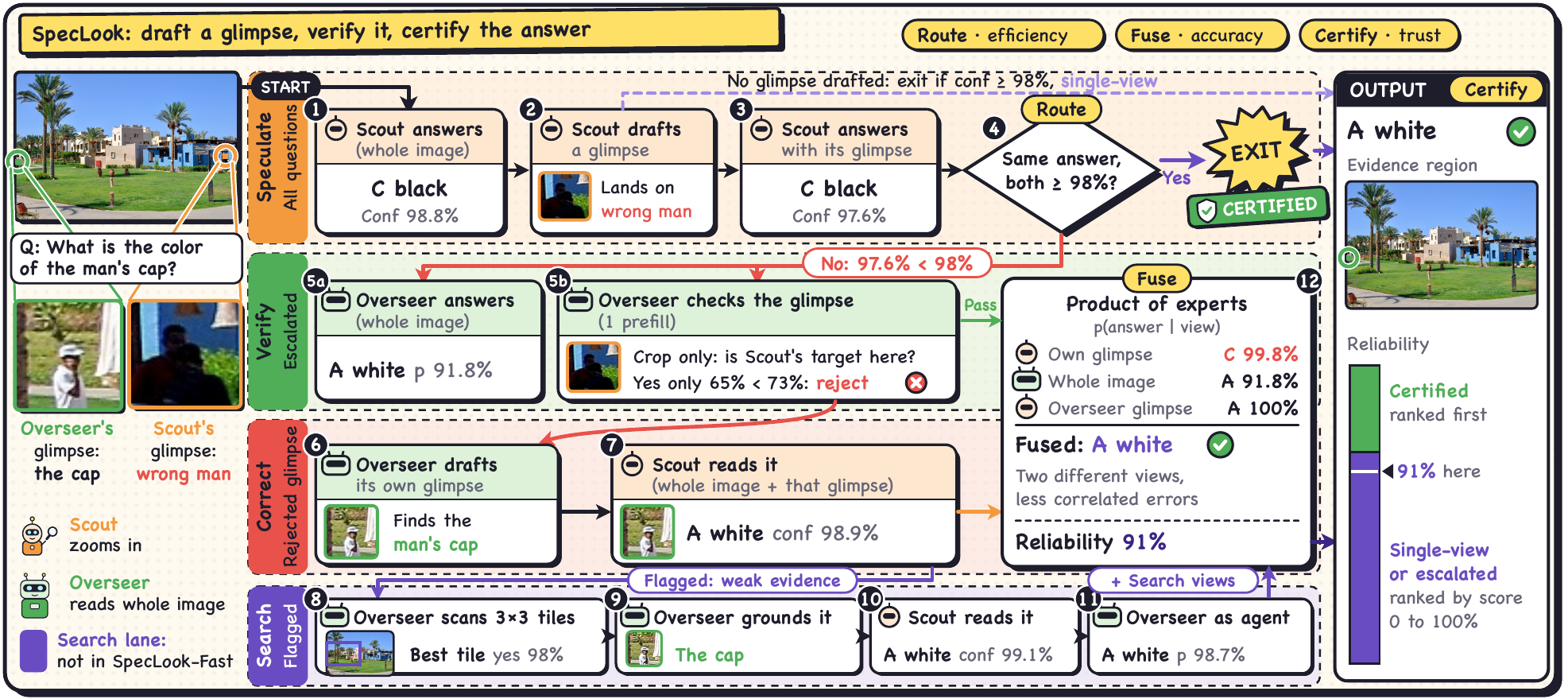}
  \caption{Overview of \method. Every question runs steps 1 to 4 and exits if the answer is unchanged and confident. Otherwise, steps 5 to 7 verify and correct the glimpse, steps 8 to 11 search flagged questions, and step 12 fuses the views. \methodfast skips steps 8 to 11.}
  \label{fig:framework}
\end{figure*}

\method is a training-free speculate-and-verify framework in which a small \scout drafts a glimpse of the evidence region and a large MLLM, the \overseer, verifies it (\cref{fig:framework}).

\subsection{Problem Formulation}

Let $I$ be an image and $q$ a question about it.
An agentic MLLM $G$ answers by alternating reasoning and tool calls~\cite{yao2023react}. At step $t$ it generates reasoning text, emits a box $b_t$, receives the crop $I[b_t]$ as a new observation, and encodes it.
We seek a training-free procedure $\pi$ that returns an answer $\hat{a}$, the evidence region $e$ it was read from, and a reliability score $r$:
\begin{equation}
  \min_{\pi} \ T(\pi) \quad \text{s.t.} \quad \mathrm{Acc}(\pi) \geq \mathrm{Acc}(G),
  \label{eq:objective}
\end{equation}
where $T$ is the mean time per question, $\mathrm{Acc}$ the accuracy on each benchmark, and $r$ should rank correct answers above wrong ones with no labels.
\method assigns new roles to the two models of SpecEyes~\cite{huang2026speceyes}. The \scout $S$ is the small model (Qwen3-VL-2B), and the \overseer $L$ is the large model of the agentic MLLM $G$ (DeepEyes-7B or Thyme-RL).
For a model $m \in \{S, L\}$ and an input view $v$, we write $a_m(v)$ for its greedy answer and $c_m(v) \in [0, 1]$ for its answer confidence, the separability score of SpecEyes.
We write $p_m(a \mid v)$ for the distribution over the candidate answers $\mathcal{A}$ of $q$, from the probabilities of the first generated token renormalized over $\mathcal{A}$, and $\ell_m(v) = \log p_m(\text{yes} \mid v) - \log p_m(\text{no} \mid v)$ for the yes/no log-odds.

\subsection{Speculate: Drafting the Glimpse}

The \scout makes two short calls on every question and a third if it drafts a glimpse.
First, it answers from the whole image, giving the draft answer $a_0 = a_S(I)$ with confidence $c_0 = c_S(I)$.
Second, without answering, it drafts up to three boxes, each with an optional text label, or an empty list when the whole image is needed.
The glimpse $g$ is the bounding box of the union of these boxes.
Third, a drafted glimpse is executed as a crop, and the \scout re-answers from the closer view $(I, I[g])$, the whole image followed by the crop, giving $a_1 = a_S(I, I[g])$ with confidence $c_1 = c_S(I, I[g])$.
A small region is upsampled, so its crop gives the target more visual tokens than the whole image does.
The trace of a question holds $(a_0, c_0, g, a_1, c_1)$ and the outputs of the \overseer after an escalation.

\subsection{Route: Certified Exit}

A confident wrong draft passes any threshold on the confidence of its own view.
The closer view adds a second test at the cost of one short call. If the draft is correct and its evidence is in the glimpse, the answer should be unchanged after zooming.
\method accepts the draft only then:
\begin{equation}
  \text{exit} \iff a_0 = a_1 \ \text{and} \ \min(c_0, c_1) \geq \tau,
  \label{eq:exit}
\end{equation}
and returns $a_1$.
We call this a certified exit, as the draft has passed an empirical test on new evidence, and \cref{sec:analysis} measures the error rate of this rule.
When the \scout drafts no valid glimpse, the question exits if $c_0 \geq \tau$, which we call a single-view exit.
For the same \scout and $\tau$, every question that exits under these two rules also exits under the confidence rule $c_0 \geq \tau$, with the same answer, so the accepted errors of \method are a subset of those of the confidence rule, and the second test can only remove errors.

\subsection{Verify: Check and Correction}

An escalated question goes through at most three \overseer stages, each batched across questions.
First, the \overseer answers from the whole image alone, which gives $p_L(a \mid I)$.
Second, it verifies the glimpse. For each label $t$, it receives the crop $I[g]$ alone with the question ``Is there a $\langle t \rangle$ in this image?'', and its first token gives the log-odds $\ell_L(I[g], t)$.
The glimpse is rejected if $\min_t \ell_L(I[g], t) < \delta$, that is, if the crop has a ``yes'' probability below $\operatorname{sigmoid}(\delta)$ for some label (73\% at $\delta = 1$).
A glimpse without labels is kept.
Third, after a rejection, the \overseer drafts a corrective glimpse $g'$ from the whole image, and if $g'$ is non-empty the \scout answers from $(I, I[g'])$.
This extends the draft-and-verify pattern of speculative decoding to the zoom step, with a drafted glimpse in place of a drafted token.
A question thus costs at most three \overseer stages and four \scout calls, even after a rejection of the glimpse.

\subsection{Fuse: Product of Experts}

The escalated answer combines the (model, view) pairs in the trace.
Let $\mathcal{V}$ contain the \scout on its closer view $(I, I[g])$ (or on $I$ when no glimpse was drafted), the \overseer on the whole image $I$, and, after a correction, the \scout on $(I, I[g'])$.
When the glimpse output does not follow the box format, its complete boxes form a recovered glimpse that replaces $g$ on escalated questions.
The fused answer is
\begin{equation}
  a^\star = \arg\max_{a \in \mathcal{A}} \sum_{(m, v) \in \mathcal{V}} \log p_m(a \mid v).
  \label{eq:fusion}
\end{equation}
This product of experts~\cite{hinton2002training}, the product rule of classifier combination~\cite{kittler1998combining}, is the Bayes decision under a uniform prior when the views are conditionally independent given the answer and each $p_m(\cdot \mid v)$ is the posterior of its view.

\textbf{Why the \overseer keeps the whole image.}
The view assignment is chosen by the oracle accuracy of the pair, an upper bound on any rule that selects one of two answers.

\begin{proposition}
\label{prop:joint_error}
Let the answers of the \scout and of the \overseer have error rates $e_S, e_L \in (0, 1)$, with $\sigma_m = \sqrt{e_m(1 - e_m)}$ and error correlation $\phi$.
Any rule that returns one of the two answers has accuracy at most $1 - e_S e_L - \phi\, \sigma_S \sigma_L$.
Replacing the \overseer answer by one with a lower error rate $e_{L'} < e_L$ and correlation $\phi'$ lowers this bound if and only if $\phi' \sigma_S \sigma_{L'} - \phi\, \sigma_S \sigma_L > e_S (e_L - e_{L'})$.
\end{proposition}

The proof is in the supplementary material.
The bound is the probability that at least one of the two answers is correct, as no such rule is correct when both are wrong.
By the proposition, a view for the \overseer preserves this bound only if $e_S$ times the accuracy it adds is at least the increase in the error covariance $\phi\, \sigma_S \sigma_L$, both measurable from the two models' answers.
A view that makes the \overseer more accurate alone can thus lower the oracle accuracy of the pair.
If the \overseer also answered with the glimpse, a misleading crop would affect both answers and $\phi$ would rise.
Only the \scout therefore answers from the closer view, while the \overseer keeps the whole image, which a misleading crop cannot affect.

\textbf{Presence questions.}
Every yes/no question in our benchmarks asks whether an object is present.
A whole-image view misses small objects, while a zoomed view tends to report absent ones.
An object is present if it is visible in some region, so \method pools the $K \leq 3$ views of the \scout as a union: the whole image, the glimpse view or a recovered glimpse, and the corrected glimpse view.
With the \overseer on $I$, the answer is yes if and only if
\begin{equation}
  \max_{k} \ell_k - \log \max\!\left(1, \tfrac{K}{2}\right) + \ell_L(I) > 0,
  \label{eq:presence}
\end{equation}
where $\ell_k$ is the log-odds of the \scout on its $k$-th view, and the $\log(K/2)$ term keeps the union bound on false positives at its two-view value when a third view is added.

\subsection{Certify: Reliability from the Trace}

Every answer has an evidence region $e$ and a reliability score $r$ from the trace, with no extra call and no labels:
\begin{equation}
  r =
  \begin{cases}
    1 + \min(c_0, c_1) & \text{certified exit,} \\
    p_S(a_0 \mid I) & \text{single-view exit,} \\
    \bar{p}(a^\star) & \text{escalated,}
  \end{cases}
  \label{eq:reliability}
\end{equation}
where $\bar{p}$ is the normalized pool of the $n = |\mathcal{V}|$ pairs,
\begin{equation}
  \bar{p}(a) \propto \exp\Big(\frac{1}{n} \sum_{(m, v) \in \mathcal{V}} \log p_m(a \mid v)\Big).
  \label{eq:normalized_pool}
\end{equation}
The pool has the same maximizers as \cref{eq:fusion}.
Certified exits score above 1 and rank first, and a single-view exit scores the pool of its one view, the \scout on $I$.
For a yes/no question, $\bar{p}$ scores the union-rule answer.
With two or more views, the score uses their agreement, which no single-view confidence has.
It ranks the answers, as selective prediction requires~\cite{geifman2017selective}, and the reliability metrics use only its order.
The evidence region $e$ is the last region the \scout read: $g$, a recovered or corrective glimpse, or the whole image $I$.

\subsection{Search: Evidence for Flagged Questions}
\label{sec:agent}

\method spends one bounded evidence search on the questions whose trace shows weak evidence.
It searches a multiple-choice question that is neither a certified exit nor answered from a glimpse that passed the check, and an escalated yes/no question when $r < \tau$.
The \overseer names the targets and checks the $3 \times 3$ overlapping tiles for them.
For multiple-choice questions, the \scout reads the best tile if its score is at least $\delta$, the \overseer grounds the targets in it with up to three boxes, and the \scout reads the region of these boxes if it passes the check.
The agent then answers in one forward pass in its own tool format, with the glimpse regions and the best tile given as its earlier tool calls and its first answer token read without decoding.
This is the only agent call of \method, and no question runs the agent loop.
\Cref{eq:fusion} then fuses the \scout readings whose region passes the check (all if none passes) and one \overseer vote, split equally between its whole-image answer and the agent answer.
For yes/no questions, \cref{eq:presence} adds $\max_j \ell_L(I_j) - \log \tfrac{9}{2}$ over the tiles $I_j$.
The search adds at most seven \overseer stages, each batched across questions, and two \scout calls to each searched question.

\begin{table*}[t]
  \centering
  \caption{Main results. Acc. in \%. s/q: seconds per question. Spd.: speedup over the agent, green for methods that accelerate it (deeper: faster). All splits: mean Acc. over the splits, s/q and Spd. over all questions. Bold: best (Spd.: green cells). Underline: second best.}
  \label{tab:main}
  \small
  \setlength{\tabcolsep}{3.2pt}
  \setlength{\aboverulesep}{0pt}\setlength{\belowrulesep}{0pt}\setlength{\extrarowheight}{0.5pt}
  \begin{tabular}{l cc cc cc cc cc cc | ccc}
  \toprule
  \rule[-0.7ex]{0pt}{0pt} & \multicolumn{2}{c}{\multirow{2}{*}{V*}} & \multicolumn{4}{c}{HR-Bench} & \multicolumn{6}{c|}{POPE} & \multicolumn{3}{c}{\multirow{2}{*}{All splits}} \\
  \cmidrule(lr){4-7} \cmidrule(l){8-13}
  \rule[-0.7ex]{0pt}{0pt} & \multicolumn{2}{c}{} & \multicolumn{2}{c}{4K} & \multicolumn{2}{c}{8K} & \multicolumn{2}{c}{Rand.} & \multicolumn{2}{c}{Pop.} & \multicolumn{2}{c|}{Adv.} & \multicolumn{3}{c}{} \\
  \cmidrule(lr){2-3} \cmidrule(lr){4-5} \cmidrule(lr){6-7} \cmidrule(lr){8-9} \cmidrule(lr){10-11} \cmidrule(l){12-13} \cmidrule(r){14-16}
  \rule[-0.7ex]{0pt}{0pt}Method & Acc. & Spd. & Acc. & Spd. & Acc. & Spd. & Acc. & Spd. & Acc. & Spd. & Acc. & Spd. & Acc. & s/q $\downarrow$ & Spd. $\uparrow$ \\
  \midrule
  \rowcolor{grouprow} \multicolumn{13}{l|}{\textit{DeepEyes-7B as the expert}} & \multicolumn{3}{l}{} \\
  DeepEyes-7B agent & 89.0 & 1.0$\times$ & 74.6 & 1.0$\times$ & 70.8 & 1.0$\times$ & 92.1 & 1.0$\times$ & 85.5 & 1.0$\times$ & 81.8 & 1.0$\times$ & 82.3 & 9.92 & 1.0$\times$ \\
  DeepEyes-7B w/o tools & 78.0 & 5.3$\times$ & 75.9 & 2.4$\times$ & 72.6 & 2.7$\times$ & 89.5 & 34.9$\times$ & 88.3 & 28.0$\times$ & 87.3 & 40.4$\times$ & 81.9 & 1.28 & 7.7$\times$ \\
  Qwen3-VL-2B (draft only) & 80.6 & 6.7$\times$ & 71.5 & 3.5$\times$ & 67.8 & 2.4$\times$ & 92.1 & 37.2$\times$ & 89.1 & 30.2$\times$ & 87.6 & 36.3$\times$ & 81.5 & 1.20 & 8.3$\times$ \\
  ViCrop & 57.6 & 2.6$\times$ & 53.9 & 1.8$\times$ & 46.4 & 0.7$\times$ & 90.3 & 13.5$\times$ & 88.4 & 10.8$\times$ & 87.1 & 13.2$\times$ & 70.6 & 3.62 & 2.7$\times$ \\
  \softrule
  ZoomEye & 86.9 & 0.8$\times$ & 73.0 & 0.5$\times$ & 72.5 & 0.3$\times$ & 87.3 & \spd{25}{3.3$\times$} & 86.5 & \spd{15}{2.5$\times$} & 85.8 & \spd{25}{3.2$\times$} & 82.0 & 10.03 & 1.0$\times$ \\
  SpecReason & 58.6 & 0.6$\times$ & 55.6 & 0.4$\times$ & 49.8 & 0.5$\times$ & 68.3 & 0.4$\times$ & 61.4 & 0.4$\times$ & 59.0 & 0.5$\times$ & 58.8 & 21.76 & 0.5$\times$ \\
  Spec. decoding & 88.0 & 0.8$\times$ & 74.8 & 1.0$\times$ & 70.5 & 0.8$\times$ & 92.2 & 0.8$\times$ & 85.6 & 1.2$\times$ & 82.0 & 0.8$\times$ & 82.2 & 11.39 & 0.9$\times$ \\
  SpecEyes & 89.5 & \spd{10}{1.8$\times$} & 74.4 & \spd{5}{1.4$\times$} & 70.5 & 1.1$\times$ & 92.7 & \spd{20}{2.8$\times$} & 89.5 & \spd{10}{1.8$\times$} & 87.7 & \spd{10}{2.1$\times$} & 84.1 & \spd{10}{5.57} & \spd{10}{1.8$\times$} \\
  \rowcolor{oursrow} \textbf{\methodfast} & \second{91.6} & \spd{50}{\textbf{5.4$\times$}} & \second{79.9} & \spd{65}{\textbf{6.7$\times$}} & \second{75.8} & \spd{55}{\textbf{6.0$\times$}} & \second{92.8} & \spd{100}{\textbf{14.1$\times$}} & \second{90.4} & \spd{100}{\textbf{13.2$\times$}} & \second{88.8} & \spd{100}{\textbf{12.4$\times$}} & \second{86.5} & \spd{100}{\textbf{1.00}} & \spd{100}{\textbf{9.9$\times$}} \\
  \rowcolor{oursrow} \textbf{\method} & \textbf{92.7} & \spd{30}{3.9$\times$} & \textbf{81.0} & \spd{35}{4.3$\times$} & \textbf{76.8} & \spd{30}{3.8$\times$} & \textbf{93.1} & \spd{100}{13.7$\times$} & \textbf{90.6} & \spd{100}{12.5$\times$} & \textbf{89.0} & \spd{100}{11.8$\times$} & \textbf{87.2} & \spd{75}{1.28} & \spd{75}{7.7$\times$} \\
  \midrule
  \rowcolor{grouprow} \multicolumn{13}{l|}{\textit{Thyme-RL as the expert}} & \multicolumn{3}{l}{} \\
  Thyme-RL agent & 84.3 & 1.0$\times$ & 77.9 & 1.0$\times$ & 72.5 & 1.0$\times$ & 91.3 & 1.0$\times$ & 86.3 & 1.0$\times$ & 82.7 & 1.0$\times$ & 82.5 & 4.95 & 1.0$\times$ \\
  Thyme-RL w/o tools & 81.2 & 2.7$\times$ & 74.5 & 1.1$\times$ & 71.1 & 1.4$\times$ & 89.2 & 20.6$\times$ & 87.9 & 10.4$\times$ & 87.4 & 12.5$\times$ & 81.9 & 1.63 & 3.0$\times$ \\
  Qwen3-VL-2B (draft only) & 80.6 & 4.4$\times$ & 71.5 & 2.0$\times$ & 67.8 & 1.6$\times$ & 92.1 & 21.3$\times$ & 89.1 & 12.5$\times$ & 87.6 & 12.7$\times$ & 81.5 & 1.20 & 4.1$\times$ \\
  ViCrop & 58.1 & 2.0$\times$ & 54.1 & 0.7$\times$ & 46.4 & 0.5$\times$ & 90.8 & 7.7$\times$ & 88.9 & 4.5$\times$ & 87.3 & 4.6$\times$ & 70.9 & 3.54 & 1.4$\times$ \\
  \softrule
  ZoomEye & 83.8 & 0.4$\times$ & 72.9 & 0.2$\times$ & 71.3 & 0.1$\times$ & 86.0 & \spd{10}{1.9$\times$} & 85.3 & 1.0$\times$ & 84.9 & 1.0$\times$ & 80.7 & 12.77 & 0.4$\times$ \\
  SpecReason & 83.8 & 0.6$\times$ & 60.8 & 0.3$\times$ & 54.6 & 0.4$\times$ & 85.8 & 0.5$\times$ & 80.9 & 0.3$\times$ & 77.8 & 0.3$\times$ & 73.9 & 14.23 & 0.3$\times$ \\
  Spec. decoding & 84.3 & 0.6$\times$ & 78.1 & 0.6$\times$ & 71.3 & 0.5$\times$ & 91.0 & 1.0$\times$ & 86.7 & 0.4$\times$ & 82.3 & 0.5$\times$ & 82.3 & 8.57 & 0.6$\times$ \\
  SpecEyes & 85.3 & \spd{5}{1.3$\times$} & 78.5 & 0.7$\times$ & 73.4 & 0.8$\times$ & 92.4 & \spd{5}{1.6$\times$} & 89.5 & \spd{10}{1.7$\times$} & 87.3 & 1.2$\times$ & 84.4 & 4.42 & 1.1$\times$ \\
  \rowcolor{oursrow} \textbf{\methodfast} & \second{89.5} & \spd{30}{\textbf{3.9$\times$}} & \second{78.6} & \spd{40}{\textbf{4.5$\times$}} & \second{74.4} & \spd{40}{\textbf{4.5$\times$}} & \second{92.6} & \spd{85}{\textbf{8.7$\times$}} & \second{89.8} & \spd{80}{\textbf{8.1$\times$}} & \second{88.3} & \spd{80}{\textbf{8.0$\times$}} & \second{85.5} & \spd{60}{\textbf{0.78}} & \spd{60}{\textbf{6.3$\times$}} \\
  \rowcolor{oursrow} \textbf{\method} & \textbf{90.1} & \spd{20}{2.7$\times$} & \textbf{80.0} & \spd{20}{2.8$\times$} & \textbf{75.4} & \spd{20}{2.8$\times$} & \textbf{92.8} & \spd{80}{8.4$\times$} & \textbf{90.1} & \spd{75}{7.6$\times$} & \textbf{88.6} & \spd{70}{7.4$\times$} & \textbf{86.2} & \spd{40}{1.05} & \spd{40}{4.7$\times$} \\
  \bottomrule
  \end{tabular}
\end{table*}

\section{Experiments}
\label{sec:experiments}

\subsection{Experimental Setup}
\label{sec:setup}

\textbf{Benchmarks and models.}
V*~\cite{wu2024vstar} (191 questions) and HR-Bench 4K and 8K~\cite{wang2025divide} (800 each) ask multiple-choice questions on small objects in high-resolution images, and POPE~\cite{li2023pope} asks if an object is present, with 3,000 questions in each of its random, popular, and adversarial splits.
Benchmarks and models follow SpecEyes~\cite{huang2026speceyes}. The \scout is Qwen3-VL-2B~\cite{bai2025qwen3vl}, and the \overseer is DeepEyes-7B~\cite{zheng2026deepeyes} or Thyme-RL~\cite{zhang2026thyme}.
We call these two large models the experts and their tool-using versions the agents.

\textbf{Baselines.}
We compare with each expert's agent, the expert without tools (w/o tools), the \scout alone (draft only), ViCrop~\cite{zhang2025mllms}, ZoomEye~\cite{shen2025zoomeye}, SpecReason~\cite{pan2025specreason}, speculative decoding~\cite{leviathan2023fast} of the agent (Spec. decoding) with a Qwen2.5-VL-3B~\cite{bai2025qwen25vl} draft, and SpecEyes.
Every baseline runs its released code and batch size with the same models.

\textbf{Metrics.}
Accuracy follows the released SpecEyes scoring rules on V* and HR-Bench and the POPE authors' protocol, with unanswered questions counted as wrong.
Speed is seconds per question of online inference after model loading and the speedup over the same expert's agent.
Reliability is the AUROC of each score for detecting wrong answers and its selective accuracy versus coverage~\cite{geifman2017selective}.

\textbf{Implementation details.}
All runs use one NVIDIA H100 NVL GPU.
The \scout and the \overseer stay loaded in BF16, and every call decodes greedily.
We follow SpecEyes and run both versions at batch size 6 on V*, 2 on HR-Bench, and 8 on POPE in \cref{tab:main}, and at 1 elsewhere.
One setting serves every benchmark and expert: $\tau = 98\%$, $\delta = 1$, equal view weights, and up to three boxes per glimpse.
In \cref{tab:main}, \method is timed end to end, with \methodfast as the same run without its search, a separate stage.

\subsection{Main Experiments}

\begin{figure}[t]
  \centering
  \includegraphics[width=\linewidth]{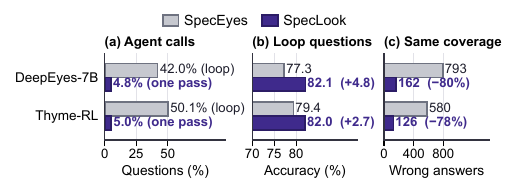}
  \caption{\method and SpecEyes on the same questions. (a) Questions that call the agent. (b) Mean accuracy where SpecEyes runs the agent loop. (c) Wrong answers among the answers SpecEyes accepts and as many top-scored answers of \method.}
  \label{fig:route_comparison}
\end{figure}

\begin{figure}[t]
  \centering
  \includegraphics[width=\linewidth]{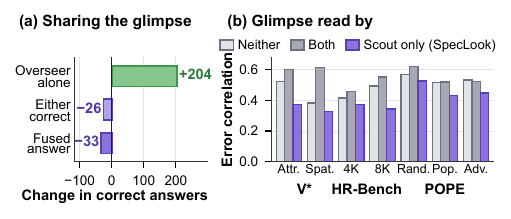}
  \caption{Giving the glimpse to the \overseer as well, with DeepEyes-7B on the 6,294 questions with a glimpse. (a) Change in correct answers. (b) Error correlation of the two models.}
  \label{fig:view_assignment}
\end{figure}

\Cref{tab:main} compares \method and \methodfast with all baselines.
Both are more accurate than every baseline on every split with either expert, and only they are both more accurate and faster than the agent on every split.
\method is the most accurate on every split and raises the mean accuracy of the DeepEyes-7B agent from 82.3\% to 87.2\% at 7.7$\times$ its speed, and \methodfast is the fastest method that accelerates the agent, at 9.9$\times$.
The same holds with Thyme-RL, which zooms by executing image-processing code.
The gains hold whether tools help the expert, as on V*, or hurt it, as on POPE popular and adversarial, and both versions are more accurate than the expert without tools and the \scout alone on every split.
Over all images, the gains in mean accuracy of both versions over the agent and SpecEyes are significant with both experts (supplementary Section S7.9).

\begin{figure*}[t]
  \centering
  \begin{minipage}[t]{0.645\linewidth}
    \vspace{0pt}
    \centering
    \includegraphics[height=56mm]{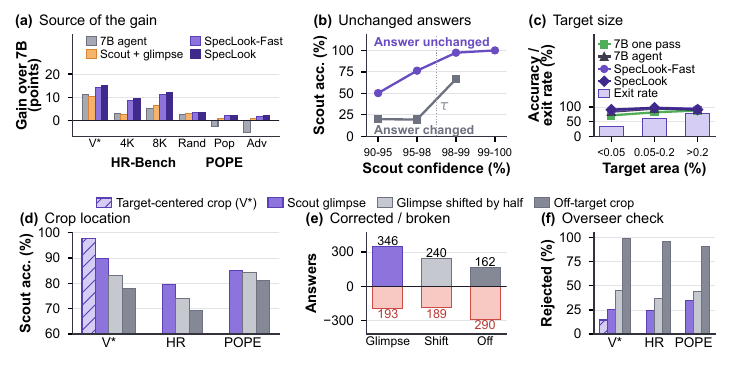}
    \caption{Why the glimpse works, with DeepEyes-7B. (a) Gain over one whole-image pass of the 7B. (b) \scout accuracy when the answer is unchanged or changed after the glimpse. (c) V* by target size. (d) to (f) The glimpse against crops of its size.}
    \label{fig:glimpse_analysis}
  \end{minipage}\hfill
  \begin{minipage}[t]{0.34\linewidth}
    \vspace{0pt}
    \centering
    \includegraphics[height=56mm]{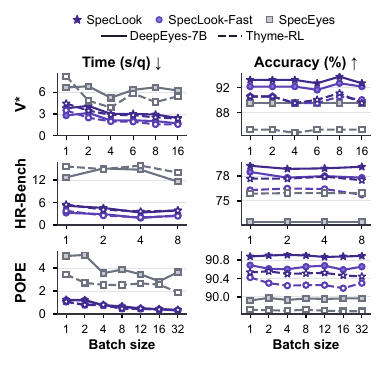}
    \caption{Speed and accuracy versus batch size.}
    \label{fig:batch_sweep}
  \end{minipage}
\end{figure*}

\Cref{fig:route_comparison} compares \method with SpecEyes on the same questions.
SpecEyes runs the full agent loop on every draft it rejects, and with Thyme-RL it is slower than the agent on both HR-Bench splits.
\method never runs the loop. Its only agent call is one forward pass on at most 5.0\% of the questions (panel a), and where SpecEyes runs the loop, \method is more accurate without it, by 4.8 and 2.7 points (panel b).
More than half of the questions reach the \overseer, and the speedup comes from a bounded cost of about four short calls per question, most by the 2B \scout, with each answer read from its first token.
The trace also ranks answers better. Among as many answers as SpecEyes accepts, those that \method scores highest hold about one fifth as many errors with either expert (panel c).

\begin{table}[t]
  \centering
  \caption{Ablation. \methodfast is \method without the search, and each variant below removes or replaces one part of \methodfast. Accuracy in \%, with HR and POPE averaged over their splits. Confidence exit: the draft is accepted on its whole-image confidence alone. Shared view: the \overseer also answers from the glimpse. Err.: wrong answers, of 10,791, returned without an \overseer call. s/q: seconds per question, averaged over both experts. Bold: best. Underline: second best. Batch size 1.}
  \label{tab:ablation}
  \footnotesize
  \setlength{\tabcolsep}{1.7pt}
  \begin{tabular}{l ccc ccc cc}
  \toprule
   & \multicolumn{3}{c}{DeepEyes-7B} & \multicolumn{3}{c}{Thyme-RL} & & \\
  \cmidrule(lr){2-4} \cmidrule(lr){5-7}
  Variant & V* & HR & POPE & V* & HR & POPE & Err. & s/q $\downarrow$ \\
  \midrule
  \rowcolor{oursrow} \textbf{\method} & \textbf{93.2} & \textbf{79.3} & \textbf{90.9} & \textbf{90.6} & \textbf{77.7} & \textbf{90.5} & 234 & 1.78 \\
  \rowcolor{oursrow} \textbf{\methodfast} & 92.1 & 78.5 & \second{90.7} & \textbf{90.6} & 76.3 & \second{90.4} & 240 & 1.48 \\
  \midrule
  \rowcolor{grouprow} \multicolumn{9}{l}{\textit{Speculate}} \\
  w/o glimpse & 81.7 & 71.5 & 89.6 & 82.2 & 70.2 & 89.2 & 700 & 0.22 \\
  \midrule
  \rowcolor{grouprow} \multicolumn{9}{l}{\textit{Route}} \\
  confidence exit & 90.1 & \second{78.6} & 90.1 & 88.5 & 76.3 & 90.0 & 700 & 1.35 \\
  w/o early exit & 92.1 & 78.5 & \second{90.7} & \textbf{90.6} & 76.3 & \second{90.4} & 0 & 1.57 \\
  \midrule
  \rowcolor{grouprow} \multicolumn{9}{l}{\textit{Verify}} \\
  w/o check \& correction & 90.6 & 78.3 & 90.6 & \second{89.5} & 76.3 & \second{90.4} & 240 & 1.35 \\
  \midrule
  \rowcolor{grouprow} \multicolumn{9}{l}{\textit{Fuse}} \\
  \overseer only & 82.2 & 70.2 & 88.6 & 83.8 & 69.4 & 88.0 & 240 & 1.48 \\
  shared view & \second{92.7} & 76.9 & 90.3 & \second{89.5} & \second{77.3} & 90.3 & 240 & 1.50 \\
  \bottomrule
  \end{tabular}
\end{table}

\subsection{Ablation Study}

\Cref{tab:ablation} removes the search from \method and then one part of \methodfast at a time.
The variants include the common form of each part: no glimpse (answer speculation), a confidence exit (cascades), an unverified crop, and a view shared by both models (ensembles).
The glimpse matters most. Without it, V* accuracy with DeepEyes-7B falls by 10.4 points, and the \overseer alone is far below the fused answer on V* and HR-Bench.
Removing the exits changes no column, so they spare the \overseer at no accuracy cost.
With the confidence exit of SpecEyes on the same outputs, 700 wrong answers skip the \overseer instead of 240, and V* accuracy falls with both experts.
The wrong exits of \methodfast are all single-view exits, which the trace ranks below every certified exit.
RePOPE~\cite{neuhaus2025repope} corrects or removes the POPE label of most of them, and on this route the agent loop is less accurate than the exits (supplementary Sections S7.8 and S7.10).
The check and correction raise V* accuracy with both experts, and removing them raises no column.
With a shared view and DeepEyes-7B, the fused answer is correct on fewer questions although the \overseer alone is correct on more, as the errors of the two models are more correlated (\cref{fig:view_assignment}, \cref{prop:joint_error}).
A single model reading its own glimpse (2B or 7B) is less accurate than \methodfast on every split, and answer speculation with a \scout zoom before its gate has a lower mean accuracy and runs the agent loop on 53.6\% of the questions.

\begin{figure*}[t]
  \centering
  \includegraphics[width=\linewidth]{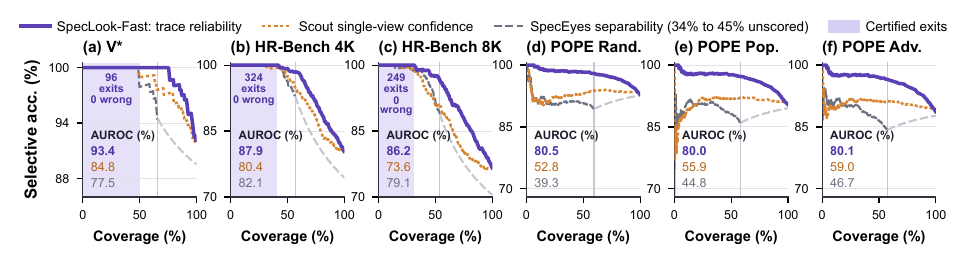}
  \caption{Selective accuracy of \methodfast versus coverage with DeepEyes-7B at batch size 1.}
  \label{fig:trust}
\end{figure*}

\textbf{Search.}
The search gives the best accuracy in every column for 0.3 seconds more per question.
It changes answers only on the flagged questions, where accuracy rises from 60.1\% to 63.0\% with DeepEyes-7B, while the agent loop there reaches 55.0\% at about half the speedup of \method.
With a stronger agent in the search, SenseNova-MARS-8B~\cite{chng2025sensenovamars}, HR-Bench accuracy rises further on both splits with both experts (supplementary Section S5).

\textbf{Batch size and \scout size.}
Each stage is one short call per question, so a batch of questions advances together. With DeepEyes-7B on POPE, raising the batch size from 1 to 32 cuts the \methodfast time about fourfold at nearly the same accuracy (\cref{fig:batch_sweep}), and both versions are faster than SpecEyes at every batch size with both experts.
A larger \scout, Qwen3-VL-4B, keeps \methodfast ahead of the DeepEyes-7B agent on every split and raises its speedup at batch size 1 to 7.1$\times$ and its accuracy to 92.7\% on V* and 81.4\% on HR-Bench 4K (supplementary Table S6).

\subsection{Analysis}
\label{sec:analysis}

\textbf{Where the gain comes from.}
The agent loop gains most in its first look. With one glimpse, the 2B \scout recovers most of the agent's gain over one whole-image pass of its 7B, leading the agent on four of six splits and staying close to it on the others (\cref{fig:glimpse_analysis}a).
Fusing the whole-image view of the 7B adds to this gain on every split, and a misleading crop is weighed against that view.
On POPE popular and adversarial, the loop leaves the agent below its own single pass of the 7B, while both versions stay above it.

\textbf{Answer stability as a test.}
Whether an answer survives zooming carries information that confidence does not. At similar \scout confidence, drafts whose answer is unchanged after zooming are far more often correct than drafts whose answer changes (\cref{fig:glimpse_analysis}b), 76.4\% against 19.6\% just below the exit threshold on V* and HR-Bench, which is why the certified exit asks for both.
The exit also adapts to difficulty without any estimate of target size. On V*, the exit rate more than doubles from the smallest targets to the largest, so the \overseer is spent where the target is hardest to see, and there both versions stay ahead of the agent (\cref{fig:glimpse_analysis}c).

\textbf{Where the glimpse looks.}
What matters is where the glimpse looks, not that a second view is added (\cref{fig:glimpse_analysis}d to f).
Crops of the same size lower the \scout accuracy the further they move from the glimpse, and on V* a crop centered on the annotated target raises it.
The glimpse corrects nearly twice as many wrong drafts as it turns into errors. A crop shifted by half still corrects more than it turns into errors, and an off-target crop turns more than it corrects, so the gain fades gradually as the crop drifts off the target.
The \overseer check follows the same order. It rejects a crop more often the further it lies from the annotated target, and at least nine in ten off-target crops, so it verifies where the \scout looked, much as speculative decoding verifies drafted tokens.
With DeepEyes-7B, its corrections more than double the glimpses that contain all annotated target centers, and the final evidence region contains them on 149 of the 191 V* questions, against 123 for the agent's own zoom crops.
With a target-centered crop, the \scout is 8.1 points more accurate on V* than with its own glimpse, so a better choice of where to look can raise the accuracy further.

\textbf{Reliability.}
The trace orders answers by the evidence behind them. All 681 certified exits, on 182 images, are correct, single-view exits are less accurate, and escalated answers least.
The 680 certified exits at the \cref{tab:main} batch sizes are also correct, and no wrong answer that SpecEyes accepts with DeepEyes-7B is a certified exit.
The trace score detects wrong answers better than the SpecEyes acceptance score and the \scout's single-view confidence on every split (\cref{fig:trust}) and with both experts. At 70\% coverage, its selective accuracy on V* is 100.0\% with both experts.
With DeepEyes-7B, the margin is largest on POPE, although an absent object has no region to zoom into.
On escalated questions, the score detects errors better than the \overseer's confidence, as it pools both models' views.
Without its \scout views, the mean AUROC falls 9.6 points with DeepEyes-7B, and without its \overseer view by 0.5.
\FloatBarrier

\section{Conclusion}
\label{sec:conclusion}

We presented \method, a training-free speculate-and-verify framework that accelerates agentic MLLMs by speculating the agent's next observation instead of its final answer.
A glimpse drafted by the \scout replaces the agent loop, and its trace routes each question, gives the two models different views to fuse, scores the reliability of every answer, and flags the questions with weak evidence for a bounded search.
With one setting for all benchmarks and both experts, \method reaches the highest accuracy on all six splits of V*, HR-Bench 4K and 8K, and POPE at 7.7$\times$ and 4.7$\times$ the speed of the DeepEyes-7B and Thyme-RL agents, \methodfast is ahead of every baseline at 9.9$\times$ and 6.3$\times$ with no agent call, and no certified exit is wrong.
Two lessons carry over. Draft the step that adds evidence, here where to look, and give two combined models different evidence.
Future work includes open-ended answers and tasks with several observations of the image in sequence.
\par

{\small
    \bibliographystyle{ieeenat_fullname}
    \bibliography{references}
}

\setcounter{section}{0}
\setcounter{figure}{0}
\setcounter{table}{0}
\setcounter{equation}{0}
\setcounter{proposition}{0}
\renewcommand{\thesection}{S\arabic{section}}
\renewcommand{\thefigure}{S\arabic{figure}}
\renewcommand{\thetable}{S\arabic{table}}
\renewcommand{\theequation}{S\arabic{equation}}
\renewcommand{\theproposition}{S\arabic{proposition}}
\renewcommand{\theHsection}{S\arabic{section}}
\renewcommand{\theHfigure}{S\arabic{figure}}
\renewcommand{\theHtable}{S\arabic{table}}
\renewcommand{\theHequation}{S\arabic{equation}}
\renewcommand{\theHproposition}{S\arabic{proposition}}
\clearpage
\newpage
\twocolumn[\centering\Large\vspace*{9pt}\textbf{Supplementary Material}\\\vspace{1.0em}\vspace{16.2pt}]
\setcounter{dbltopnumber}{4}
\renewcommand{\dbltopfraction}{0.88}
\makeatletter
\setlength{\@dblfptop}{0pt}
\setlength{\@dblfpsep}{14pt}
\makeatother

This supplementary material gives the implementation details of \method and of its efficient version, \methodfast, and further results that support the main paper.
\begin{itemize}
  \setlength{\itemsep}{0pt}
  \item \cref{sec:suppl_impl}: views, prompts, answer confidence, glimpse parsing, the evidence search, timing, evaluation, and baselines.
  \item \cref{sec:suppl_results}: results with Thyme-RL, at other batch sizes, per step and per route of \methodfast, by POPE error type, and with a larger \scout.
  \item \cref{sec:suppl_ablation}: ablations of implementation details, sensitivity to the fixed setting and the exit threshold, the view assignment, single-model zoom baselines, answer speculation with one zoom, glimpse rounds, and the relation to existing techniques.
  \item \cref{sec:suppl_reliability}: further reliability metrics with confidence intervals, calibration, the parts of the trace score, other scores, the reliability on POPE, and the agent loop on the questions with the lowest trace score.
  \item \cref{sec:suppl_agent}: ablations and reliability of the evidence search of \method, its accuracy on the searched questions, and a stronger agent in the search.
  \item \cref{sec:suppl_qualitative}: a speculation trace, and questions that \method answers correctly where SpecEyes and the agent do not.
  \item \cref{sec:suppl_analysis}: formal analysis of routing, fusion, the reliability score, and cost, with proofs, the empirical error bound of the certified exits, significance tests of the accuracy gains, and POPE with corrected labels.
\end{itemize}
In result tables, bold marks the best values, underlining marks the second best, and deeper green indicates larger speedup.

\section{Implementation Details}
\label{sec:suppl_impl}

This section expands on the implementation details in Section 4.1 of the main paper.
One setting serves every benchmark and both \overseer models: $\tau = 98\%$, $\delta = 1$, which is a ``yes'' probability of $\operatorname{sigmoid}(1) \approx 73\%$ in the glimpse check, equal weights for all views, and at most three boxes per glimpse.
The threshold $\tau$ and the answer confidence follow the released SpecEyes code.
The same setting is used unchanged with the larger \scout in \cref{sec:suppl_results}, and \cref{tab:suppl_hparam} varies its values.

\textbf{Models and decoding.}
Both models stay loaded on one NVIDIA H100 NVL GPU in BF16 precision with FlashAttention-2.
With both models loaded and the search included, the peak GPU memory allocated by the runs of \method in Table 1 of the main paper is 21.4 to 28.5 GiB.
Every call uses greedy decoding, with at most 20 new tokens for an answer of the \scout, 512 for an answer of the \overseer, 160 for a glimpse, and 1 for the check of a glimpse.

\textbf{Views.}
Each image enters a model at a fixed size, so the view sets the number of visual tokens.
The whole image is resized without upsampling to at most 4,096 visual tokens, where one token covers $32 \times 32$ pixels for the \scout and $28 \times 28$ pixels for the \overseer.
Each box of a glimpse is widened to at least 56 pixels per side, and the glimpse is the bounding box of their union.
Its crop has the same cap of 4,096 tokens and is upsampled to about 256 tokens when the region is smaller.
A closer view $(I, I[g])$ consists of the whole image, a short text that gives the position of the crop, and the crop.
For the check, the \overseer receives the crop alone at the same size.

\begin{table*}[t]
  \centering
  \caption{Prompts of \method.}
  \label{tab:suppl_prompts}
  \frenchspacing
  \begin{promptbox}{Answer}{\scout, \overseer}
  \ttfamily System: You are a helpful assistant. Output only the final answer itself. Do not include explanations, reasoning, labels, or any extra text.\newline
  User: \textrm{$\langle$whole image$\rangle$} [\textrm{$\langle$crop text$\rangle$} \textrm{$\langle$crop$\rangle$}] Question: \{question\}\newline
  Options: \{options\}\newline
  Answer:
  \end{promptbox}
  \begin{promptbox}{Crop text}{}
  \ttfamily Additional evidence crop. Region [\{x1\}, \{y1\}, \{x2\}, \{y2\}] in the displayed full image (\{W\}x\{H\}). This crop is a partial observation.
  \end{promptbox}
  \begin{promptbox}{Glimpse}{\scout}
  \ttfamily System: You are a helpful assistant.\newline
  User: \textrm{$\langle$whole image$\rangle$} The image is displayed at \{W\}x\{H\} pixels.\newline
  Question: \{question with options\}\newline
  Locate the visual evidence needed to answer this question. Do not answer the question.\newline
  Output only a JSON list of zero to three objects, each containing "bbox\_2d": [x1, y1, x2, y2]. Coordinates are normalized to [0, 1000] relative to the displayed image width and height.\newline
  Include all regions needed to compare entities or read their labels and units. Keep each region's surrounding context. Order the boxes by importance.\newline
  Use [] when the complete image is needed, or no specific magnification is needed. Do not select only a small part for an exhaustive global count. Do not output transcriptions or explanations.
  \end{promptbox}
  \begin{promptbox}{Corrective glimpse}{\overseer}
  Same as Glimpse, with the coordinate sentence replaced by \texttt{Coordinates are pixels in the displayed image.}
  \end{promptbox}
  \begin{promptbox}{Check}{\overseer}
  \ttfamily System: as in Answer.\newline
  User: \textrm{$\langle$crop$\rangle$} Question: Is there a \{label\} in this image?\newline
  Answer:
  \end{promptbox}
  \begin{promptbox}{Object names}{\overseer}
  \ttfamily System: You are a helpful assistant.\newline
  User: \textrm{$\langle$whole image$\rangle$} Question: \{question\}\newline
  Which objects must be located in the image to answer this question? Output a JSON list of one or two short noun phrases that name each object by its category only, without the color, text, number or other property the question asks about, for example ["car"] or ["cup", "laptop"]. Do not answer the question.
  \end{promptbox}
  \begin{promptbox}{Tile glimpse}{\overseer}
  Corrective glimpse on the selected tile in place of the whole image.
  \end{promptbox}
  \begin{promptbox}{Agent answer}{\overseer}
  Released prompt of the agent. Each searched region follows as a zoom call of the agent, and its tool response holds the crop and the check result: \texttt{Check: the \{label\} is visible in this region.} or \texttt{Check: no \{label\} was found in this region. Zoom in on another region to find it.} The answer is read from the first token after the answer tag.
  \end{promptbox}
\end{table*}

\textbf{Prompts.}
\cref{tab:suppl_prompts} lists every prompt.
The last three calls belong to the evidence search, which \methodfast skips.
Text in braces is filled per question, and POPE questions have no options line.
Angle brackets mark images, and the crop text and crop in square brackets appear only in a closer view.
Answer prompts use the question format of the released SpecEyes code.
The glimpse prompt requests boxes, and the \scout may add a \texttt{label} field to each box in its grounding format.
A label names a target $t$ for the check, and a glimpse without labels is kept unchecked.
At batch size 1, all 6,294 valid glimpses carry a label, and the \overseer checks every one on an escalated question.
The \overseer drafts its corrective glimpse with the same prompt in pixel coordinates of its displayed image, the box format of Qwen2.5-VL.

\textbf{Answer confidence and distribution.}
As in SpecEyes, the confidence of an answer with generated tokens $1, \dots, J$ is
\begin{equation}
  c = \operatorname{sigmoid}\Big(0.6 \min_{1 \leq j \leq J} \frac{z_j - \mu_j}{\sigma_j}\Big),
\end{equation}
where $z_j$ is the largest logit at step $j$, and $\mu_j$ and $\sigma_j$ are the mean and the standard deviation of the logits ranked 2 to 64.
The distribution $p_m(a \mid v)$ renormalizes the probabilities of the first generated token over the option letters, or over ``yes'' and ``no'' with upper and lower case merged.
The confidence $c_m(v)$ routes the question, and the distribution $p_m(\cdot \mid v)$ enters fusion.

\textbf{Glimpse parsing.}
A glimpse output is valid if it is a JSON list of at most three boxes whose coordinates lie in $[0, 1000]$ relative to the displayed image.
An empty list means that the whole image is needed.
On an escalated question whose glimpse output does not follow this format, the recovered glimpse is the union of every complete box $[x_1, y_1, x_2, y_2]$ in $[0, 1000]$ in the output text, which covers outputs with more than three boxes, unclosed code fences, or bare coordinate lists.
In the HR-Bench runs of Table 1 of the main paper, the \scout reads a recovered glimpse on 8.6\% of the questions.

\textbf{Evidence search.}
\method runs its bounded evidence search on the questions that the trace flags, with the models, views, and check of the steps before it.
For a multiple-choice question, the \overseer names one or two target objects by category in a short generation, without the property that the question asks about.
For a POPE question, the target is the object in the question.
The tiles are the $3 \times 3$ windows of half the image width and height at strides of a quarter of the image, each shown at no more than 1,024 visual tokens in the check.
A tile scores the lowest check log-odds over the targets, and the best tile has the highest score.
The \scout reads the best tile when its score is at least $\delta$.
The \overseer then grounds the targets in the best tile with up to three boxes, and the \scout reads the grounded region when it passes the check.
When a draft exits with a single view, the \overseer also answers from the whole image, since no \overseer call was made for it.
The agent answer uses the released prompt of each agent.
With DeepEyes-7B, the glimpse regions and the best tile appear as earlier zoom-in calls with their crops and a note for each target that fails the check.
With Thyme-RL, they appear as earlier cropping-code turns with their outputs.
The answer is read from the option probabilities right after the answer tag, with no decoding.
\method runs at the released batch sizes, and each search stage is batched across the questions of a batch.

\textbf{Timing.}
Seconds per question cover online inference after model loading, that is, image decoding, every model call, routing, fusion, and writing the outputs, divided by the number of questions.
In Table 1 of the main paper, each split has one run of \method on the compute node of the agent run of that split.
The time of \method is the time of this run, and the time of \methodfast is the same time without the search, which is a separate stage of the run.
Table 2 of the main paper and the ablations in \cref{sec:suppl_ablation} are instead measured at batch size 1.
There, every call of \methodfast is made on every question, including the calls that an exit skips.
A variant answers from the calls it uses, and its model seconds per question are the image decoding time plus the time of these calls.

\textbf{Evaluation.}
Every answer of \method and \methodfast is an option letter or yes/no, so the released scoring rules grade it directly.
Baseline answers that these rules cannot parse are graded by Qwen2.5-72B-Instruct with the released judge prompt of SpecEyes.
We serve the judge with tensor parallel size 2 on two H100 NVL GPUs instead of the released 4.
V* has 115 attribute and 76 spatial questions, which some tables and figures report separately as Attr.\ and Spatial.
Figure 6 of the main paper also shows them separately.

\textbf{Baselines.}
Every baseline uses the expert of its block in Table 1 of the main paper, the same questions, and the scoring above.
The agents, SpecEyes, and SpecReason run the code released with SpecEyes, which provides SpecReason for both experts, and the last two use the \scout model as their small model.
SpecEyes accepts a draft above the thresholds of its released results, 98\% on V* and 97\% on HR-Bench and POPE with DeepEyes-7B, and 98\%, 98\%, and 97\% with Thyme-RL.
The expert without tools and the \scout alone answer once from the whole image.
ZoomEye runs its released tree search with its released search settings and the expert as its model.
SpecReason and ZoomEye answer one question at a time, as released.

\textbf{ViCrop and speculative decoding.}
ViCrop runs the relative-attention variant of the authors' Qwen2.5-VL code with the expert as the model, at batch size 1 and the image size of that code, at most 200,704 pixels.
At this size, the attention cells of an HR-Bench image are wider than the 224-pixel base box, so the released box rule counts zero cells and places the box at the top-left corner.
We count at least one cell, which leaves every V* and POPE box unchanged and places the HR-Bench boxes on the attention peak.
Speculative decoding runs the released agent loop of SpecEyes at batch size 1 with assisted generation in Hugging Face Transformers.
Qwen2.5-VL-3B-Instruct drafts two tokens per step, and the expert verifies every token greedily.
Its answer is that of the agent on 97.3\% of the questions with DeepEyes-7B, and its mean accuracy is within 0.2 points of the agent with both experts.
Verification scores several tokens in one forward pass, so in BF16 a nearly tied token can differ from the agent run, and the later turns of the loop then differ.

\section{Additional Results}
\label{sec:suppl_results}

\begin{figure}[t]
  \centering
  \includegraphics[width=\linewidth]{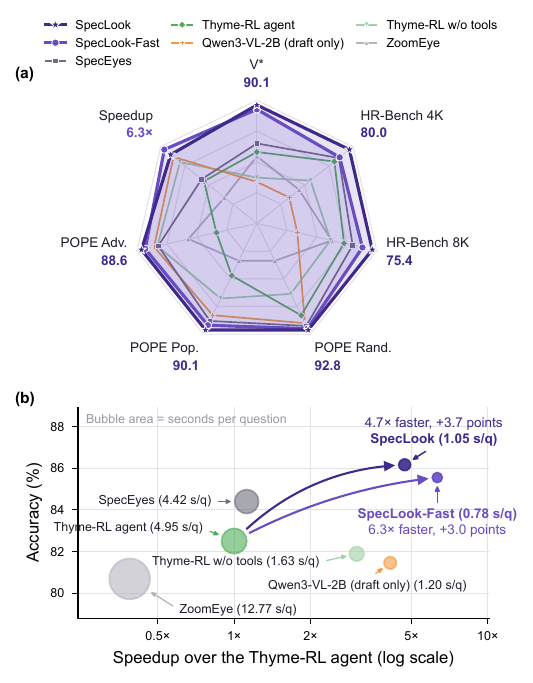}
  \caption{Accuracy and speed with Thyme-RL as the expert.}
  \label{fig:suppl_teaser_thyme}
\end{figure}

\begin{table*}[t]
  \centering
  \caption{Comparison at batch size 1. Speedup is over SpecEyes.}
  \label{tab:suppl_batch1}
  \small
  \setlength{\tabcolsep}{6pt}
  \begin{tabular}{l cccccc cc}
  \toprule
   & & \multicolumn{2}{c}{HR-Bench} & \multicolumn{3}{c}{POPE} & \multicolumn{2}{c}{All splits} \\
  \cmidrule(lr){3-4} \cmidrule(lr){5-7} \cmidrule(l){8-9}
  Method & V* & 4K & 8K & Rand. & Pop. & Adv. & s/q $\downarrow$ & Speedup $\uparrow$ \\
  \midrule
  \rowcolor{grouprow} \multicolumn{9}{l}{\textit{DeepEyes-7B as the expert, batch size 1}} \\
  ZoomEye & 86.9 & 73.0 & 72.5 & 87.3 & 86.5 & 85.8 & 10.03 & 0.6$\times$ \\
  SpecEyes & 89.5 & 74.3 & 70.6 & 92.6 & 89.5 & 87.6 & 6.20 & 1.0$\times$ \\
  \rowcolor{oursrow} \textbf{\methodfast} & \second{92.1} & \second{80.4} & \second{76.6} & \second{92.9} & \second{90.5} & \second{88.7} & \spd{35}{\textbf{1.54}} & \spd{35}{\textbf{4.0$\times$}} \\
  \rowcolor{oursrow} \textbf{\method} & \textbf{93.2} & \textbf{81.0} & \textbf{77.5} & \textbf{93.1} & \textbf{90.6} & \textbf{89.0} & \spd{25}{1.86} & \spd{25}{3.3$\times$} \\
  \midrule
  \rowcolor{grouprow} \multicolumn{9}{l}{\textit{Thyme-RL as the expert, batch size 1}} \\
  ZoomEye & 83.8 & 72.9 & 71.3 & 86.0 & 85.3 & 84.9 & 12.77 & 0.4$\times$ \\
  SpecEyes & \second{85.3} & 78.4 & 73.4 & 92.4 & 89.5 & 87.3 & 5.33 & 1.0$\times$ \\
  \rowcolor{oursrow} \textbf{\methodfast} & \textbf{90.6} & \second{78.6} & \second{74.0} & \second{92.7} & \second{90.0} & \second{88.6} & \spd{30}{\textbf{1.50}} & \spd{30}{\textbf{3.6$\times$}} \\
  \rowcolor{oursrow} \textbf{\method} & \textbf{90.6} & \textbf{80.0} & \textbf{75.4} & \textbf{92.8} & \textbf{90.2} & \textbf{88.7} & \spd{20}{1.79} & \spd{20}{3.0$\times$} \\
  \bottomrule
  \end{tabular}
\end{table*}

\begin{table*}[t]
  \centering
  \caption{Accuracy (\%) and seconds per question (s/q) versus batch size on V* and POPE.}
  \label{tab:suppl_sweep}
  \footnotesize
  \setlength{\tabcolsep}{4pt}
  \begin{minipage}[t]{0.36\linewidth}
  \centering
  \vspace{0pt}
  \begin{tabular}{l c c c}
  \toprule
  Method & Batch & Acc. & s/q \\
  \midrule
  \rowcolor{grouprow} \multicolumn{4}{l}{\textit{DeepEyes-7B as the expert}} \\
  SpecEyes & 1 & 89.5 & 6.67 \\
  SpecEyes & 2 & 89.5 & 6.85 \\
  SpecEyes & 4 & 89.5 & 5.25 \\
  SpecEyes & 6 & 89.5 & 6.35 \\
  SpecEyes & 8 & 89.5 & 6.70 \\
  SpecEyes & 16 & 89.5 & 6.26 \\
  \rowcolor{oursrow} \methodfast & 1 & 92.1 & 2.80 \\
  \rowcolor{oursrow} \methodfast & 2 & 92.1 & 3.15 \\
  \rowcolor{oursrow} \methodfast & 4 & 92.1 & 2.11 \\
  \rowcolor{oursrow} \methodfast & 6 & 91.6 & 2.14 \\
  \rowcolor{oursrow} \methodfast & 8 & 92.7 & 2.02 \\
  \rowcolor{oursrow} \methodfast & 16 & 92.1 & 1.71 \\
  \rowcolor{oursrow} \method & 1 & 93.2 & 3.79 \\
  \rowcolor{oursrow} \method & 2 & 93.2 & 4.13 \\
  \rowcolor{oursrow} \method & 4 & 93.2 & 3.02 \\
  \rowcolor{oursrow} \method & 6 & 92.7 & 3.02 \\
  \rowcolor{oursrow} \method & 8 & 93.7 & 2.94 \\
  \rowcolor{oursrow} \method & 16 & 92.7 & 2.45 \\
  \midrule
  \rowcolor{grouprow} \multicolumn{4}{l}{\textit{Thyme-RL as the expert}} \\
  SpecEyes & 1 & 85.3 & 8.24 \\
  SpecEyes & 2 & 85.3 & 4.83 \\
  SpecEyes & 4 & 84.8 & 3.91 \\
  SpecEyes & 6 & 85.3 & 5.84 \\
  SpecEyes & 8 & 85.3 & 4.66 \\
  SpecEyes & 16 & 85.3 & 5.45 \\
  \rowcolor{oursrow} \methodfast & 1 & 90.6 & 3.51 \\
  \rowcolor{oursrow} \methodfast & 2 & 90.6 & 2.49 \\
  \rowcolor{oursrow} \methodfast & 4 & 89.5 & 1.98 \\
  \rowcolor{oursrow} \methodfast & 6 & 89.5 & 1.93 \\
  \rowcolor{oursrow} \methodfast & 8 & 90.6 & 1.57 \\
  \rowcolor{oursrow} \methodfast & 16 & 90.1 & 1.58 \\
  \rowcolor{oursrow} \method & 1 & 90.6 & 4.46 \\
  \rowcolor{oursrow} \method & 2 & 90.6 & 3.46 \\
  \rowcolor{oursrow} \method & 4 & 89.5 & 2.90 \\
  \rowcolor{oursrow} \method & 6 & 90.1 & 2.84 \\
  \rowcolor{oursrow} \method & 8 & 91.1 & 2.53 \\
  \rowcolor{oursrow} \method & 16 & 89.5 & 2.37 \\
  \bottomrule
  \end{tabular}
  \end{minipage}\hfill
  \begin{minipage}[t]{0.5\linewidth}
  \centering
  \vspace{0pt}
  \begin{tabular}{l c ccc c}
  \toprule
  Method & Batch & Rand. & Pop. & Adv. & s/q \\
  \midrule
  \rowcolor{grouprow} \multicolumn{6}{l}{\textit{DeepEyes-7B as the expert}} \\
  SpecEyes & 1 & 92.6 & 89.5 & 87.6 & 5.03 \\
  SpecEyes & 2 & 92.7 & 89.6 & 87.7 & 5.13 \\
  SpecEyes & 4 & 92.6 & 89.5 & 87.7 & 3.54 \\
  SpecEyes & 8 & 92.7 & 89.5 & 87.7 & 3.86 \\
  SpecEyes & 12 & 92.6 & 89.5 & 87.7 & 3.45 \\
  SpecEyes & 16 & 92.7 & 89.5 & 87.7 & 2.84 \\
  SpecEyes & 32 & 92.7 & 89.5 & 87.7 & 3.63 \\
  \rowcolor{oursrow} \methodfast & 1 & 92.9 & 90.5 & 88.7 & 1.21 \\
  \rowcolor{oursrow} \methodfast & 2 & 92.8 & 90.4 & 88.7 & 1.19 \\
  \rowcolor{oursrow} \methodfast & 4 & 92.8 & 90.4 & 88.6 & 0.75 \\
  \rowcolor{oursrow} \methodfast & 8 & 92.8 & 90.4 & 88.8 & 0.64 \\
  \rowcolor{oursrow} \methodfast & 12 & 92.9 & 90.4 & 88.7 & 0.46 \\
  \rowcolor{oursrow} \methodfast & 16 & 92.8 & 90.3 & 88.7 & 0.41 \\
  \rowcolor{oursrow} \methodfast & 32 & 92.8 & 90.4 & 88.7 & 0.30 \\
  \rowcolor{oursrow} \method & 1 & 93.1 & 90.6 & 89.0 & 1.24 \\
  \rowcolor{oursrow} \method & 2 & 93.1 & 90.6 & 89.0 & 1.21 \\
  \rowcolor{oursrow} \method & 4 & 93.1 & 90.6 & 89.0 & 0.77 \\
  \rowcolor{oursrow} \method & 8 & 93.1 & 90.6 & 89.0 & 0.67 \\
  \rowcolor{oursrow} \method & 12 & 93.0 & 90.6 & 89.0 & 0.49 \\
  \rowcolor{oursrow} \method & 16 & 93.1 & 90.6 & 89.0 & 0.43 \\
  \rowcolor{oursrow} \method & 32 & 93.1 & 90.6 & 89.0 & 0.33 \\
  \midrule
  \rowcolor{grouprow} \multicolumn{6}{l}{\textit{Thyme-RL as the expert}} \\
  SpecEyes & 1 & 92.4 & 89.5 & 87.3 & 3.44 \\
  SpecEyes & 2 & 92.4 & 89.5 & 87.3 & 2.71 \\
  SpecEyes & 4 & 92.3 & 89.5 & 87.3 & 2.53 \\
  SpecEyes & 8 & 92.4 & 89.5 & 87.3 & 2.52 \\
  SpecEyes & 12 & 92.3 & 89.5 & 87.3 & 2.66 \\
  SpecEyes & 16 & 92.3 & 89.5 & 87.3 & 2.56 \\
  SpecEyes & 32 & 92.3 & 89.5 & 87.3 & 1.88 \\
  \rowcolor{oursrow} \methodfast & 1 & 92.7 & 90.0 & 88.6 & 1.06 \\
  \rowcolor{oursrow} \methodfast & 2 & 92.6 & 89.9 & 88.4 & 0.78 \\
  \rowcolor{oursrow} \methodfast & 4 & 92.6 & 89.8 & 88.3 & 0.78 \\
  \rowcolor{oursrow} \methodfast & 8 & 92.6 & 89.8 & 88.3 & 0.45 \\
  \rowcolor{oursrow} \methodfast & 12 & 92.6 & 89.8 & 88.3 & 0.46 \\
  \rowcolor{oursrow} \methodfast & 16 & 92.5 & 89.8 & 88.3 & 0.38 \\
  \rowcolor{oursrow} \methodfast & 32 & 92.6 & 89.8 & 88.4 & 0.34 \\
  \rowcolor{oursrow} \method & 1 & 92.8 & 90.2 & 88.7 & 1.09 \\
  \rowcolor{oursrow} \method & 2 & 92.8 & 90.2 & 88.7 & 0.81 \\
  \rowcolor{oursrow} \method & 4 & 92.8 & 90.1 & 88.6 & 0.80 \\
  \rowcolor{oursrow} \method & 8 & 92.8 & 90.1 & 88.6 & 0.48 \\
  \rowcolor{oursrow} \method & 12 & 92.8 & 90.2 & 88.6 & 0.48 \\
  \rowcolor{oursrow} \method & 16 & 92.7 & 90.0 & 88.7 & 0.41 \\
  \rowcolor{oursrow} \method & 32 & 92.7 & 90.0 & 88.6 & 0.37 \\
  \bottomrule
  \end{tabular}
  \end{minipage}
\end{table*}

\begin{table*}[t]
  \centering
  \caption{Seconds per question at the released batch sizes. All: all splits. $^\ddagger$Batch size 1.}
  \label{tab:suppl_efficiency}
  \small
  \setlength{\tabcolsep}{6pt}
  \begin{tabular}{l cccccc c}
  \toprule
   & & \multicolumn{2}{c}{HR-Bench} & \multicolumn{3}{c}{POPE} & \\
  \cmidrule(lr){3-4} \cmidrule(lr){5-7}
  Seconds per question $\downarrow$ & V* & 4K & 8K & Rand. & Pop. & Adv. & All \\
  \midrule
  \rowcolor{grouprow} \multicolumn{8}{l}{\textit{DeepEyes-7B as the expert}} \\
  DeepEyes-7B agent & 11.64 & 15.84 & 20.54 & 9.06 & 7.31 & 8.86 & 9.92 \\
  ZoomEye$^\ddagger$ & 13.89 & 29.92 & 70.51 & 2.72 & 2.97 & 2.74 & 10.03 \\
  SpecEyes & 6.35 & 11.07 & 19.16 & 3.25 & 4.10 & 4.23 & 5.57 \\
  \rowcolor{oursrow} \textbf{\methodfast} & \textbf{2.14} & \textbf{2.37} & \textbf{3.42} & \textbf{0.64} & \textbf{0.56} & \textbf{0.72} & \textbf{1.00} \\
  \rowcolor{oursrow} \quad speedup vs.\ agent & \spd{50}{5.4$\times$} & \spd{65}{6.7$\times$} & \spd{55}{6.0$\times$} & \spd{100}{14.1$\times$} & \spd{100}{13.2$\times$} & \spd{100}{12.4$\times$} & \spd{100}{9.9$\times$} \\
  \rowcolor{oursrow} \quad speedup vs.\ SpecEyes & \spd{20}{3.0$\times$} & \spd{40}{4.7$\times$} & \spd{50}{5.6$\times$} & \spd{45}{5.1$\times$} & \spd{70}{7.4$\times$} & \spd{55}{5.9$\times$} & \spd{50}{5.6$\times$} \\
  \rowcolor{oursrow} \textbf{\method} & 3.02 & 3.66 & 5.43 & 0.66 & 0.58 & 0.75 & 1.28 \\
  \rowcolor{oursrow} \quad speedup vs.\ agent & \spd{30}{3.9$\times$} & \spd{35}{4.3$\times$} & \spd{30}{3.8$\times$} & \spd{100}{13.7$\times$} & \spd{100}{12.5$\times$} & \spd{100}{11.8$\times$} & \spd{75}{7.7$\times$} \\
  \rowcolor{oursrow} \quad speedup vs.\ SpecEyes & \spd{10}{2.1$\times$} & \spd{20}{3.0$\times$} & \spd{30}{3.5$\times$} & \spd{45}{4.9$\times$} & \spd{65}{7.0$\times$} & \spd{50}{5.6$\times$} & \spd{35}{4.3$\times$} \\
  \midrule
  \rowcolor{grouprow} \multicolumn{8}{l}{\textit{Thyme-RL as the expert}} \\
  Thyme-RL agent & 7.57 & 8.98 & 13.57 & 5.19 & 3.02 & 3.09 & 4.95 \\
  ZoomEye$^\ddagger$ & 19.01 & 39.24 & 95.62 & 2.67 & 2.97 & 3.13 & 12.77 \\
  SpecEyes & 5.84 & 13.31 & 16.53 & 3.20 & 1.80 & 2.56 & 4.42 \\
  \rowcolor{oursrow} \textbf{\methodfast} & \textbf{1.93} & \textbf{2.00} & \textbf{3.01} & \textbf{0.60} & \textbf{0.37} & \textbf{0.39} & \textbf{0.78} \\
  \rowcolor{oursrow} \quad speedup vs.\ agent & \spd{30}{3.9$\times$} & \spd{40}{4.5$\times$} & \spd{40}{4.5$\times$} & \spd{85}{8.7$\times$} & \spd{80}{8.1$\times$} & \spd{80}{8.0$\times$} & \spd{60}{6.3$\times$} \\
  \rowcolor{oursrow} \quad speedup vs.\ SpecEyes & \spd{20}{3.0$\times$} & \spd{60}{6.6$\times$} & \spd{50}{5.5$\times$} & \spd{50}{5.3$\times$} & \spd{40}{4.8$\times$} & \spd{60}{6.6$\times$} & \spd{50}{5.6$\times$} \\
  \rowcolor{oursrow} \textbf{\method} & 2.84 & 3.23 & 4.91 & 0.62 & 0.40 & 0.42 & 1.05 \\
  \rowcolor{oursrow} \quad speedup vs.\ agent & \spd{20}{2.7$\times$} & \spd{20}{2.8$\times$} & \spd{20}{2.8$\times$} & \spd{80}{8.4$\times$} & \spd{75}{7.6$\times$} & \spd{70}{7.4$\times$} & \spd{40}{4.7$\times$} \\
  \rowcolor{oursrow} \quad speedup vs.\ SpecEyes & \spd{10}{2.1$\times$} & \spd{35}{4.1$\times$} & \spd{25}{3.4$\times$} & \spd{45}{5.2$\times$} & \spd{40}{4.5$\times$} & \spd{55}{6.1$\times$} & \spd{35}{4.2$\times$} \\
  \bottomrule
  \end{tabular}
\end{table*}

\begin{figure*}[t]
  \centering
  \includegraphics[width=\linewidth]{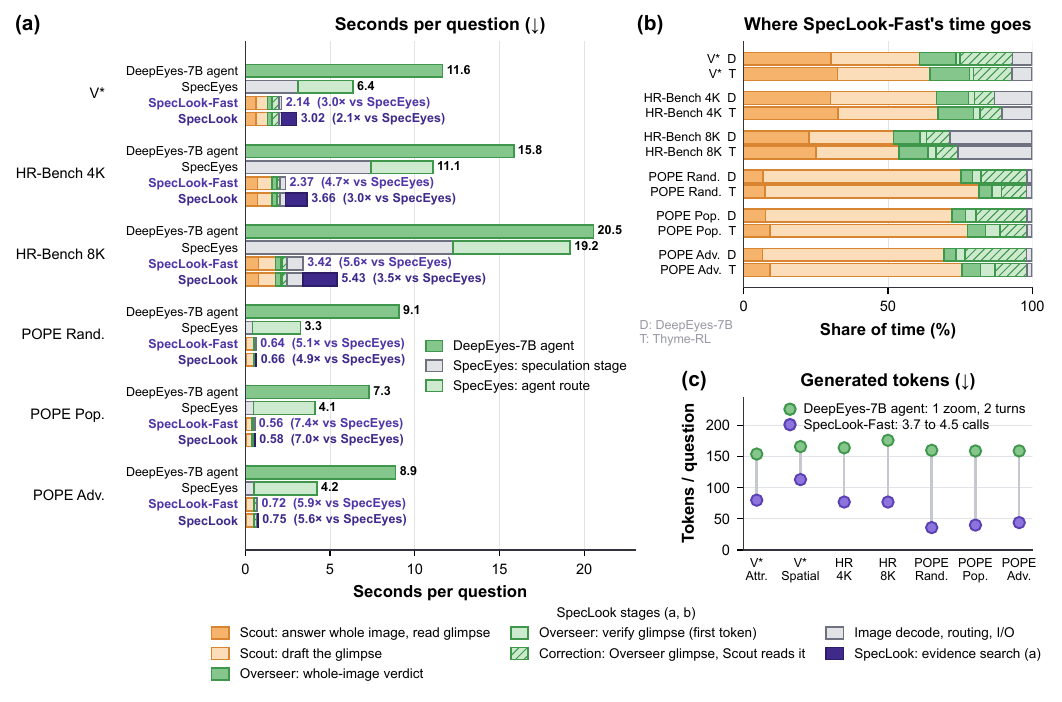}
  \caption{Time and generated tokens at the released batch sizes.}
  \label{fig:suppl_efficiency}
\end{figure*}

\begin{figure*}[t]
  \centering
  \includegraphics[width=\linewidth]{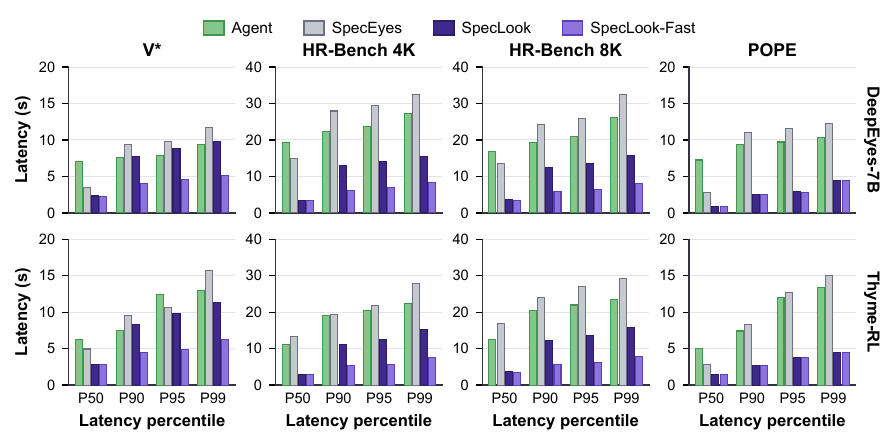}
  \caption{Seconds per question at the 50th, 90th, 95th, and 99th percentile at batch size 1.}
  \label{fig:suppl_tail_latency}
\end{figure*}

\begin{table*}[t]
  \centering
  \caption{Routes of \methodfast at batch size 1. Share: \% of the questions. Acc.: accuracy (\%) on the route.}
  \label{tab:suppl_routes}
  \small
  \setlength{\tabcolsep}{6pt}
  \begin{tabular}{l cc cc cc cc}
  \toprule
   & \multicolumn{2}{c}{V*} & \multicolumn{2}{c}{HR-Bench} & \multicolumn{2}{c}{POPE} & \multicolumn{2}{c}{All questions} \\
  \cmidrule(lr){2-3} \cmidrule(lr){4-5} \cmidrule(lr){6-7} \cmidrule(l){8-9}
  Route & Share & Acc. & Share & Acc. & Share & Acc. & Share & Acc. \\
  \midrule
  \rowcolor{grouprow} \multicolumn{9}{l}{\textit{DeepEyes-7B as the expert}} \\
  Certified exit & 50.3 & 100.0 & 35.8 & 100.0 & 0.1 & 100.0 & 6.3 & 100.0 \\
  Single-view exit & 0.0 & -- & 1.7 & 77.8 & 43.3 & 94.0 & 36.4 & 93.9 \\
  Escalated to the \overseer & 49.7 & 84.2 & 62.5 & 66.2 & 56.5 & 88.1 & 57.3 & 84.5 \\
  \midrule
  \rowcolor{grouprow} \multicolumn{9}{l}{\textit{Thyme-RL as the expert}} \\
  Certified exit & 50.3 & 100.0 & 35.8 & 100.0 & 0.1 & 100.0 & 6.3 & 100.0 \\
  Single-view exit & 0.0 & -- & 1.7 & 77.8 & 43.3 & 94.0 & 36.4 & 93.9 \\
  Escalated to the \overseer & 49.7 & 81.1 & 62.5 & 62.7 & 56.5 & 87.7 & 57.3 & 83.5 \\
  \bottomrule
  \end{tabular}
\end{table*}

\begin{figure*}[t]
  \centering
  \includegraphics[width=0.9\linewidth]{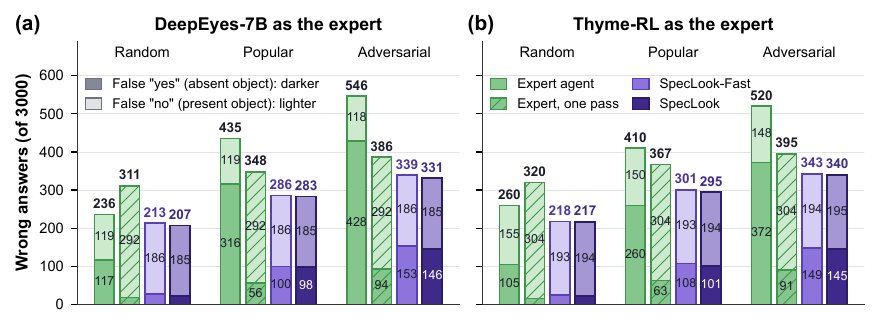}
  \caption{Wrong POPE answers by type at batch size 1. Expert, one pass: the expert answers once from the whole image.}
  \label{fig:suppl_pope_errors}
\end{figure*}

\begin{table*}[t]
  \centering
  \caption{A larger \scout with DeepEyes-7B at batch size 1. Speedup is over the agent.}
  \label{tab:suppl_scout_size}
  \small
  \setlength{\tabcolsep}{6pt}
  \begin{tabular}{l l cccccc cc}
  \toprule
   & & & \multicolumn{2}{c}{HR-Bench} & \multicolumn{3}{c}{POPE} & \multicolumn{2}{c}{All splits} \\
  \cmidrule(lr){4-5} \cmidrule(lr){6-8} \cmidrule(l){9-10}
  Method & \scout & V* & 4K & 8K & Rand. & Pop. & Adv. & s/q $\downarrow$ & Speedup $\uparrow$ \\
  \midrule
  DeepEyes-7B agent & -- & 89.0 & 74.6 & 70.8 & 92.1 & 85.5 & 81.8 & 9.92 & 1.0$\times$ \\
  \rowcolor{oursrow} \methodfast & Qwen3-VL-2B & \second{92.1} & \second{80.4} & \textbf{76.6} & \textbf{92.9} & \textbf{90.5} & \textbf{88.7} & \spd{60}{1.54} & \spd{60}{6.5$\times$} \\
  \rowcolor{oursrow} \methodfast & Qwen3-VL-4B & \textbf{92.7} & \textbf{81.4} & \second{76.4} & \second{92.3} & \second{90.0} & \second{88.5} & \spd{70}{\textbf{1.41}} & \spd{70}{\textbf{7.1$\times$}} \\
  \bottomrule
  \end{tabular}
\end{table*}

\textbf{Thyme-RL as the expert.}
\cref{fig:suppl_teaser_thyme} repeats Figure 1 of the main paper with Thyme-RL as the expert.
Panel (a) reports each split and speedup over the agent, and panel (b) plots mean accuracy over the six splits against speedup, with bubble area proportional to seconds per question.
Both versions are more accurate than every baseline on every split.
\method is the most accurate method on every split, raises the mean accuracy over the six splits from 82.5\% for the agent to 86.2\%, and is 4.7$\times$ faster than the agent.
\methodfast reaches 85.5\% and is 6.3$\times$ faster.

\textbf{Batch size.}
Table 1 of the main paper uses the released batch sizes.
\cref{tab:suppl_batch1} compares the methods with one question at a time.
Both versions remain more accurate than SpecEyes and ZoomEye on every split with both experts, and \method gives the best accuracy on every split.
\methodfast is 4.0$\times$ faster than SpecEyes with DeepEyes-7B and 3.6$\times$ faster with Thyme-RL, and \method is 3.3$\times$ and 3.0$\times$ faster.
The largest gains over SpecEyes are on HR-Bench with DeepEyes-7B, 6.1 points on 4K and 6.0 points on 8K for \methodfast and 6.7 and 6.9 points for \method.
\cref{tab:suppl_sweep} varies the batch size on V* and POPE, that is, the number of questions served at once on one GPU.
At every batch size, both versions are more accurate and faster than SpecEyes on every split with both experts.
On POPE, the time of \methodfast falls from 1.21 seconds per question at batch size 1 to 0.30 at batch size 32 with DeepEyes-7B, and from 1.06 to 0.34 with Thyme-RL.
The time of \method falls from 1.24 to 0.33 and from 1.09 to 0.37.

\textbf{Time per step.}
\cref{tab:suppl_efficiency} gives the seconds per question on each split at the released batch sizes.
\methodfast is the fastest method on every split, and it is 3.0$\times$ to 7.4$\times$ faster than SpecEyes with DeepEyes-7B and 3.0$\times$ to 6.6$\times$ faster with Thyme-RL.
\method is 2.1$\times$ to 7.0$\times$ faster than SpecEyes with DeepEyes-7B and 2.1$\times$ to 6.1$\times$ faster with Thyme-RL.
\cref{fig:suppl_efficiency} divides the time into the steps of \methodfast.
Panels (a) and (c) show seconds and generated tokens per question with DeepEyes-7B, and panel (a) adds the evidence search of \method.
Panel (b) shows each step's share of time with DeepEyes-7B (D) and Thyme-RL (T).
The short calls of the \scout take the largest share of the time on every split, and \methodfast generates 36 to 113 tokens per question against 154 to 176 for the DeepEyes-7B agent.

\textbf{Tail latency.}
\cref{fig:suppl_tail_latency} reports the seconds per question at the 50th, 90th, 95th, and 99th percentile at batch size 1.
For each split, \method, SpecEyes, and the agent run one after another on the same GPU after three warm-up questions, and every question is timed.
\methodfast is below the agent and SpecEyes at every percentile on every benchmark with both experts.
\method is below SpecEyes at every percentile on every benchmark, and below the agent at every percentile on HR-Bench and POPE.
On HR-Bench 4K with DeepEyes-7B, the 99th percentile is 8.5 seconds for \methodfast and 15.4 for \method, against 27.1 for the agent and 32.5 for SpecEyes.
SpecEyes runs the full agent loop after a rejected draft, so its 99th percentile is above the agent's on every benchmark with both experts.

\textbf{Routes.}
\cref{tab:suppl_routes} gives the share of the questions on each route and the accuracy of the answers there.
The exits depend on the \scout alone, so they are the same with both experts.
Certified exits end 50.3\% of the V* questions and 35.8\% of the HR-Bench questions, all with a correct answer.
On POPE, most exits are single-view exits, for which the \scout drafts no valid glimpse, and their answers are 94.0\% correct, against 88.1\% and 87.7\% for the escalated answers with the two experts.
In total, 42.7\% of the questions end at an exit with no \overseer call.

\textbf{POPE error types.}
\cref{fig:suppl_pope_errors} splits the POPE errors into false ``yes'' answers on absent objects and false ``no'' answers on present objects.
Darker bars mark false ``yes'', lighter bars mark false ``no'', and ``Expert, one pass'' is the \overseer answering once from the whole image.
The zoom loop of the agent adds false ``yes'' answers, and one whole-image pass of the 7B expert misses small objects.
\method makes the fewest errors on every split with both experts, followed by \methodfast.
On the adversarial split with DeepEyes-7B, \methodfast lowers the false ``yes'' answers of the agent from 428 to 153 and the false ``no'' answers of one whole-image pass from 292 to 186, and \method lowers the false ``yes'' answers further to 146.

\textbf{A larger \scout.}
\cref{tab:suppl_scout_size} replaces the \scout by Qwen3-VL-4B with the same $\tau$.
\methodfast is then more accurate than the agent on every split, reaches 92.7\% on V* and 81.4\% on HR-Bench 4K, and its speedup at batch size 1 rises from 6.5$\times$ to 7.1$\times$.

\section{Additional Ablations}
\label{sec:suppl_ablation}

\begin{figure*}[t]
  \centering
  \includegraphics[width=0.85\linewidth]{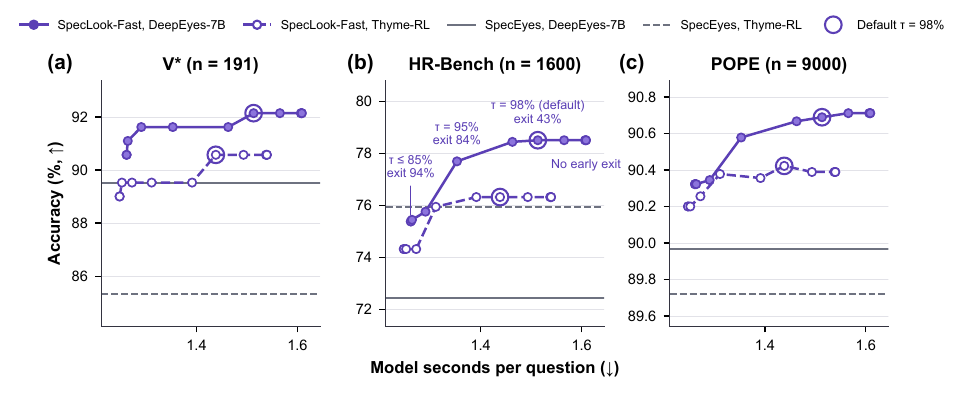}
  \caption{Effect of the exit threshold $\tau$ at batch size 1.}
  \label{fig:suppl_tau_sweep}
\end{figure*}

\begin{table*}[t]
  \begin{minipage}[t]{0.485\linewidth}
  \centering
  \caption{Ablation of implementation details at batch size 1. Accuracy in \%.}
  \label{tab:suppl_ablation_details}
  \footnotesize
  \setlength{\tabcolsep}{3pt}
  \begin{tabular}{l ccc ccc}
  \toprule
   & \multicolumn{3}{c}{DeepEyes-7B} & \multicolumn{3}{c}{Thyme-RL} \\
  \cmidrule(lr){2-4} \cmidrule(l){5-7}
  Variant & V* & HR & POPE & V* & HR & POPE \\
  \midrule
  \rowcolor{oursrow} \textbf{\methodfast} & \textbf{92.1} & \textbf{78.5} & \textbf{90.7} & \textbf{90.6} & \textbf{76.3} & \textbf{90.4} \\
  \midrule
  w/o presence union & \textbf{92.1} & \textbf{78.5} & 90.5 & \textbf{90.6} & \textbf{76.3} & \textbf{90.4} \\
  w/o count correction & \textbf{92.1} & \textbf{78.5} & 90.6 & \textbf{90.6} & \textbf{76.3} & \textbf{90.4} \\
  w/o recovered glimpses & \textbf{92.1} & 78.2 & \textbf{90.7} & \textbf{90.6} & 76.1 & \textbf{90.4} \\
  \bottomrule
  \end{tabular}
  \par\vspace{8pt}
  \expandafter\def\csname @captype\endcsname{figure}
  \includegraphics[width=0.8\linewidth]{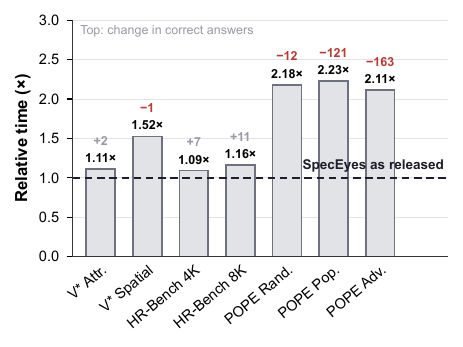}
  \caption{Estimated time of SpecEyes at the exit error rate of \methodfast, relative to its released time, with DeepEyes-7B.}
  \label{fig:suppl_exit_cost}
  \end{minipage}\hfill
  \begin{minipage}[t]{0.485\linewidth}
  \centering
  \caption{Accuracy (\%) around the fixed setting at batch size 1. Shaded: the setting of the paper.}
  \label{tab:suppl_hparam}
  \footnotesize
  \setlength{\tabcolsep}{3pt}
  \begin{tabular}{l ccc ccc}
  \toprule
   & \multicolumn{3}{c}{DeepEyes-7B} & \multicolumn{3}{c}{Thyme-RL} \\
  \cmidrule(lr){2-4} \cmidrule(l){5-7}
  Setting & V* & HR & POPE & V* & HR & POPE \\
  \midrule
  \rowcolor{grouprow} \multicolumn{7}{l}{\textit{Check threshold $\delta$ (\methodfast)}} \\
  $\delta = 0$ & 91.6 & 78.4 & 90.6 & 90.1 & 76.5 & 90.4 \\
  $\delta = 0.5$ & 91.6 & 78.4 & 90.7 & 90.6 & 76.4 & 90.4 \\
  \rowcolor{oursrow} $\delta = 1$ & 92.1 & 78.5 & 90.7 & 90.6 & 76.3 & 90.4 \\
  $\delta = 1.5$ & 92.1 & 78.5 & 90.7 & 90.6 & 76.3 & 90.4 \\
  $\delta = 2$ & 92.7 & 78.4 & 90.6 & 90.6 & 76.2 & 90.4 \\
  \midrule
  \rowcolor{grouprow} \multicolumn{7}{l}{\textit{\scout view weight (\methodfast)}} \\
  0.5 & 90.6 & 77.4 & 90.5 & 90.1 & 74.8 & 90.1 \\
  0.75 & 92.1 & 77.8 & 90.6 & 90.1 & 75.8 & 90.2 \\
  \rowcolor{oursrow} 1 & 92.1 & 78.5 & 90.7 & 90.6 & 76.3 & 90.4 \\
  1.25 & 92.1 & 77.9 & 90.5 & 91.1 & 76.7 & 90.4 \\
  1.5 & 92.7 & 77.7 & 90.5 & 90.1 & 76.7 & 90.3 \\
  2 & 92.7 & 76.9 & 90.4 & 90.6 & 76.6 & 90.2 \\
  \midrule
  \rowcolor{grouprow} \multicolumn{7}{l}{\textit{Boxes per glimpse (\methodfast)}} \\
  1 & 89.5 & 76.5 & 90.7 & 88.5 & 74.8 & 90.4 \\
  2 & 92.1 & 78.2 & 90.7 & 90.6 & 76.1 & 90.4 \\
  \rowcolor{oursrow} 3 & 92.1 & 78.5 & 90.7 & 90.6 & 76.3 & 90.4 \\
  \midrule
  \rowcolor{grouprow} \multicolumn{7}{l}{\textit{Agent share of the \overseer vote (\method)}} \\
  0.25 & 93.2 & 79.3 & 90.9 & 90.1 & 77.5 & 90.5 \\
  \rowcolor{oursrow} 0.5 & 93.2 & 79.3 & 90.9 & 90.6 & 77.7 & 90.5 \\
  0.75 & 93.2 & 79.3 & 90.9 & 90.6 & 77.9 & 90.5 \\
  \bottomrule
  \end{tabular}
  \end{minipage}
\end{table*}

\begin{table*}[t]
  \centering
  \caption{Fused accuracy and error correlation for each view assignment, with DeepEyes-7B on the 6,294 questions with a drafted glimpse.}
  \label{tab:suppl_view_assignment}
  \small
  \begin{tabular}{ll ccc}
  \toprule
  \scout sees & \overseer sees & V* & HR-Bench & POPE \\
  \midrule
  \rowcolor{grouprow} \multicolumn{5}{l}{\textit{Accuracy (\%) of the fused answer}} \\
  whole image & whole image & 80.9 & 76.5 & 86.0 \\
  whole image & whole image + glimpse & 85.6 & 82.6 & 86.5 \\
  glimpse & whole image + glimpse & 89.4 & 82.9 & 87.1 \\
  \rowcolor{oursrow} \textbf{glimpse} & \textbf{whole image} & \textbf{90.4} & \textbf{84.1} & \textbf{87.4} \\
  \midrule
  \rowcolor{grouprow} \multicolumn{5}{l}{\textit{Error correlation $\phi$ between the two models}} \\
  whole image & whole image & 0.47 & 0.46 & 0.54 \\
  whole image & whole image + glimpse & \textbf{0.28} & 0.37 & 0.53 \\
  glimpse & whole image + glimpse & 0.61 & 0.51 & 0.54 \\
  \rowcolor{oursrow} \textbf{glimpse} & \textbf{whole image} & 0.35 & \textbf{0.36} & \textbf{0.46} \\
  \bottomrule
  \end{tabular}
\end{table*}

\begin{table*}[t]
  \centering
  \caption{Single-model zoom baselines with DeepEyes-7B at batch size 1. SpecEyes: released batch sizes. Accuracy in \%. s/q: model seconds per question.}
  \label{tab:suppl_zoom_refine}
  \small
  \begin{tabular}{l cc cc ccc c}
  \toprule
   & \multicolumn{2}{c}{V*} & \multicolumn{2}{c}{HR-Bench} & \multicolumn{3}{c}{POPE} & \\
  \cmidrule(lr){2-3} \cmidrule(lr){4-5} \cmidrule(lr){6-8}
  Method & Attr. & Spatial & 4K & 8K & Rand. & Pop. & Adv. & s/q \\
  \midrule
  SpecEyes & 91.3 & \second{86.8} & 74.4 & 70.5 & \second{92.7} & \second{89.5} & \second{87.7} & 5.57 \\
  7B zoom-refine (\overseer plans and reads its glimpse) & \second{92.2} & 85.5 & \second{76.3} & 67.8 & 90.1 & 88.4 & 87.3 & 1.26 \\
  2B zoom-refine (\scout plans and reads its glimpse) & 90.4 & 85.5 & 74.3 & \second{72.0} & 92.6 & 89.1 & \second{87.7} & 1.11 \\
  \rowcolor{oursrow} \textbf{\methodfast} & \textbf{93.9} & \textbf{89.5} & \textbf{80.4} & \textbf{76.6} & \textbf{92.9} & \textbf{90.5} & \textbf{88.7} & 1.51 \\
  \bottomrule
  \end{tabular}
\end{table*}

\begin{table*}[t]
  \centering
  \caption{Answer speculation with one zoom-in step of the small model before the gate, at batch size 1. Questions that the gate does not accept run the agent loop. Accuracy in \%. Mean: mean accuracy over the six splits. Loop: \% of the questions that run the agent loop.}
  \label{tab:suppl_depth}
  \small
  \begin{tabular}{l cccc cccc c}
  \toprule
   & \multicolumn{4}{c}{DeepEyes-7B} & \multicolumn{4}{c}{Thyme-RL} & \\
  \cmidrule(lr){2-5} \cmidrule(lr){6-9}
  Method & V* & HR & POPE & Mean & V* & HR & POPE & Mean & Loop $\downarrow$ \\
  \midrule
  Depth 0, confidence gate & 89.5 & 73.1 & 89.3 & 83.9 & 85.3 & 75.5 & 89.2 & 84.0 & 23.1 \\
  Depth 1, confidence gate & \second{92.7} & 73.6 & 88.4 & 84.1 & \textbf{91.1} & \second{76.3} & 88.6 & 84.9 & 53.6 \\
  Depth 1, agreement gate & 90.1 & 73.0 & 88.3 & 83.5 & 86.9 & 75.4 & 88.5 & 83.9 & 57.3 \\
  \midrule
  \rowcolor{oursrow} \textbf{\methodfast} & 92.1 & \second{78.5} & \second{90.7} & \second{86.9} & \second{90.6} & \second{76.3} & \second{90.4} & \second{85.7} & 0.0 \\
  \rowcolor{oursrow} \textbf{\method} & \textbf{93.2} & \textbf{79.3} & \textbf{90.9} & \textbf{87.4} & \second{90.6} & \textbf{77.7} & \textbf{90.5} & \textbf{86.3} & 0.0 \\
  \bottomrule
  \end{tabular}
\end{table*}

\begin{table*}[t]
  \centering
  \caption{Additional glimpse rounds. Mean: mean accuracy over the six splits. Speedup is over the agent.}
  \label{tab:suppl_rounds}
  \small
  \setlength{\tabcolsep}{6pt}
  \begin{tabular}{l cccccc c c}
  \toprule
   & & \multicolumn{2}{c}{HR-Bench} & \multicolumn{3}{c}{POPE} & & \\
  \cmidrule(lr){3-4} \cmidrule(lr){5-7}
  Method & V* & 4K & 8K & Rand. & Pop. & Adv. & Mean & Spd. $\uparrow$ \\
  \midrule
  \rowcolor{grouprow} \multicolumn{9}{l}{\textit{DeepEyes-7B as the expert}} \\
  \rowcolor{oursrow} \methodfast (one glimpse round) & 91.6 & 79.9 & 75.8 & 92.8 & 90.4 & 88.8 & 86.5 & \spd{100}{9.9$\times$} \\
  \quad + up to two more rounds & 91.6 & 78.6 & 75.3 & 92.8 & 90.4 & 88.8 & 86.2 & \spd{80}{8.3$\times$} \\
  \rowcolor{oursrow} \method & 92.7 & 81.0 & 76.8 & 93.1 & 90.6 & 89.0 & 87.2 & \spd{75}{7.7$\times$} \\
  \midrule
  \rowcolor{grouprow} \multicolumn{9}{l}{\textit{Thyme-RL as the expert}} \\
  \rowcolor{oursrow} \methodfast (one glimpse round) & 89.5 & 78.6 & 74.4 & 92.6 & 89.8 & 88.3 & 85.5 & \spd{60}{6.3$\times$} \\
  \quad + up to two more rounds & 90.1 & 78.8 & 74.3 & 92.6 & 89.8 & 88.3 & 85.6 & \spd{50}{5.6$\times$} \\
  \rowcolor{oursrow} \method & 90.1 & 80.0 & 75.4 & 92.8 & 90.1 & 88.6 & 86.2 & \spd{40}{4.7$\times$} \\
  \bottomrule
  \end{tabular}
\end{table*}

The results in this section use the batch-1 calls of \cref{sec:suppl_impl}, as Table 2 of the main paper does.

\textbf{Implementation details.}
\cref{tab:suppl_ablation_details} removes one detail at a time.
Without the presence union, a yes/no answer adds the log-odds of the last view of the \scout to that of the \overseer instead of taking the maximum over the views of the \scout.
The row ``w/o count correction'' drops the term $\log \max(1, K/2)$.
\cref{sec:suppl_presence} derives this correction.
Without recovered glimpses, a glimpse output that does not follow the box format yields no glimpse on escalated questions as well.
With DeepEyes-7B, each detail raises the accuracy on the benchmark where it applies, POPE for the union and its count correction and HR-Bench for the recovered glimpses, and removing a detail raises no column.

\textbf{Sensitivity to the setting.}
\cref{tab:suppl_hparam} varies the fixed values one at a time.
The box cap changes the drafted glimpse, so its rows are separate runs, and the other groups use the same calls.
For $\delta$ between 0 and 2, the accuracy changes by at most one V* question and by at most 0.2 points on HR-Bench and POPE.
For a \scout weight between 0.75 and 1.25, it stays within 0.7 points of the equal weights in every column, and the equal weights give the highest HR-Bench accuracy with DeepEyes-7B.
With at most two boxes per glimpse, the accuracy is within 0.3 points of three boxes in every column.
A single box lowers V* by 2.6 and 2.1 points, all on the spatial questions, and HR-Bench from 78.5\% to 76.5\% and from 76.3\% to 74.8\%.
The share of the agent answer in the \overseer vote of \method changes only the V* and HR-Bench columns with Thyme-RL, by at most 0.5 points.

\textbf{Exit threshold.}
\cref{fig:suppl_tau_sweep} varies $\tau$ on the same batch-1 calls.
Each point is one threshold, gray lines show SpecEyes on the same questions, and large circles mark the default $\tau = 98\%$.
A lower $\tau$ lets more drafts exit and lowers the model time.
At the default $\tau = 98\%$, the accuracy equals that without the early exit in every column of Table 2 of the main paper, and 42.7\% of the questions end with no \overseer call.
With DeepEyes-7B, \methodfast is more accurate than SpecEyes at every $\tau$ on all three benchmarks.
\Cref{fig:suppl_exit_cost} gives the cost for SpecEyes of reaching the error rate of our exits with a stricter threshold.
There, SpecEyes keeps its most confident accepted answers up to the error rate of our exits on each split and sends the others to its agent loop, at the mean agent-loop time per question.
The comparison uses DeepEyes-7B, bars give time relative to SpecEyes as released, and top labels give the change in correct answers.
This takes up to 2.23$\times$ its released time and costs up to 163 correct answers on a split.

\textbf{View assignment.}
\cref{tab:suppl_view_assignment} varies which model sees which view on the 6,294 questions with a drafted glimpse, with DeepEyes-7B as the \overseer.
One answer of each model is fused by a plain product of experts, with no exit, no presence rule, and no check.
The assignment of \method, the glimpse for the \scout and the whole image for the \overseer, has the highest fused accuracy on all three benchmarks.
When the \overseer also answers from the glimpse, the error correlation rises on every benchmark, from 0.35 to 0.61 on V*.
It is then correct on 204 more questions alone, but the fused answer is correct on 33 fewer (Figure 6 of the main paper).
On paired answers, only our assignment is correct on 113 questions and only the shared view on 80.
In Table 2 of the main paper, which adds the exits, the check, and the presence rule, \method is ahead of the shared view in all six columns and \methodfast in four.

\textbf{Single-model zoom.}
In the zoom-refine baselines of \cref{tab:suppl_zoom_refine}, one model drafts a glimpse with the glimpse prompt and answers again from the whole image and its crop.
The SpecEyes row uses its released accuracy and seconds per question.
\methodfast is more accurate than both on every split.
The gain thus comes from dividing the views between the two models rather than from zooming alone.

\textbf{Answer speculation with one zoom.}
SpecEyes names multi-depth speculation as future work, in which the small model makes a bounded number of tool calls before the gate~\cite{huang2026speceyes}.
\Cref{tab:suppl_depth} builds its first depth from the same \scout calls at $\tau = 98\%$.
At depth 0, the \scout answers from the whole image, and at depth 1 it answers after its one zoom-in step.
The confidence gate accepts an answer with a confidence of at least $\tau$, the agreement gate applies the two exit rules of \method, and every other question runs the agent loop.
One zoom before the gate raises the mean accuracy of the confidence gate by 0.2 and 0.9 points, and the agent loop then runs on 53.6\% of the questions instead of 23.1\%.
\methodfast has a higher mean accuracy than every depth variant with both experts, by 2.7 and 0.8 points over the depth-1 confidence gate, and runs the agent loop on no question.
On V*, the depth-1 confidence gate is within one question of both versions with either expert, and it runs the agent loop on 31.4\% of the V* questions.
The gain of \method over answer speculation thus comes from verifying and fusing the glimpse rather than from adding a zoom before the gate.

\textbf{Glimpse rounds.}
\methodfast uses one verified glimpse per question, which matches the single zoom step of the agent.
In \cref{tab:suppl_rounds}, the \overseer proposes up to two more glimpses on the escalated questions where the two models still disagree.
The \scout reads each new glimpse.
The base runs use released batch sizes, and extra rounds are timed at batch size 1.
More rounds lower the speedup from 9.9$\times$ to 8.3$\times$ with DeepEyes-7B and from 6.3$\times$ to 5.6$\times$ with Thyme-RL, and they change the mean accuracy by at most 0.3 points.
More glimpses of the same kind therefore add little, while the evidence search of \method raises the mean accuracy to 87.2\% with DeepEyes-7B and 86.2\% with Thyme-RL (\cref{sec:suppl_agent}).

\begin{figure*}[t]
  \centering
  \includegraphics[width=0.82\linewidth]{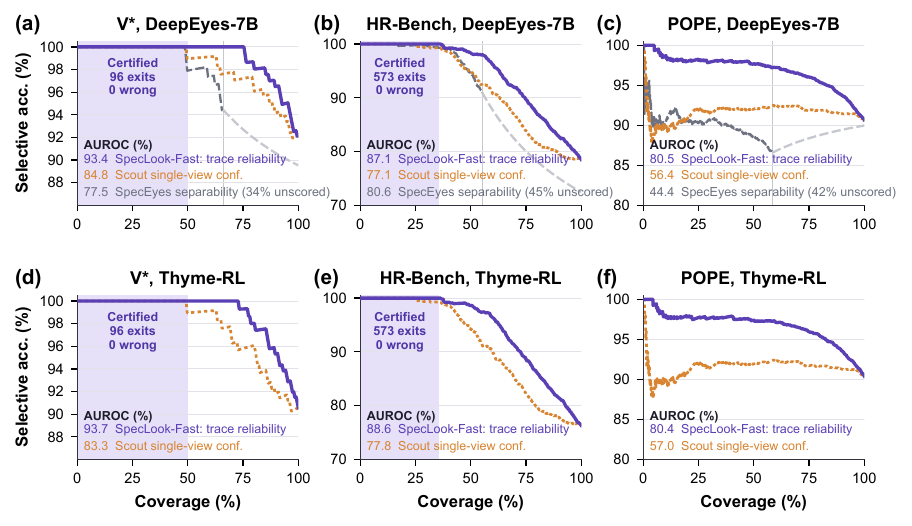}
  \caption{Selective accuracy of \methodfast versus coverage with both experts at batch size 1.}
  \label{fig:suppl_risk_coverage}
\end{figure*}

\begin{table*}[t]
  \centering
  \caption{Reliability of \methodfast at batch size 1. AUROC, ECE, and selective accuracy in \%, AURC scaled by 100. 95\% CI: bootstrap interval of the AUROC. --: the score is not defined on every answer.}
  \label{tab:suppl_trust}
  \small
  \begin{tabular}{ll l ccc c ccc}
  \toprule
   & & & & & & & \multicolumn{3}{c}{Selective acc. at coverage} \\
  \cmidrule(l){8-10}
  Expert & Bench. & Score & AUROC $\uparrow$ & 95\% CI & AURC $\downarrow$ & ECE $\downarrow$ & 50\% & 70\% & 90\% \\
  \midrule
  \rowcolor{oursrow} DeepEyes-7B & V* & \textbf{Trace reliability (ours)} & \textbf{93.4} & [88.9, 97.0] & \textbf{0.87} & \textbf{3.5} & \textbf{100.0} & \textbf{100.0} & \textbf{96.5} \\
   &  & \scout single-view conf. & 84.8 & [76.3, 91.3] & 1.70 & 15.8 & 99.0 & 97.8 & 94.2 \\
   &  & SpecEyes separability & 77.5 & [70.4, 83.8] & 3.24 & -- & 97.9 & 93.3 & 91.3 \\
  \cmidrule(l){2-10}
  \rowcolor{oursrow}  & HR & \textbf{Trace reliability (ours)} & \textbf{87.1} & [83.1, 90.5] & \textbf{5.45} & \textbf{8.1} & \textbf{98.3} & \textbf{91.9} & \textbf{82.7} \\
   &  & \scout single-view conf. & 77.1 & [71.9, 82.0] & 7.94 & 24.4 & 94.8 & 86.6 & 79.6 \\
   &  & SpecEyes separability & 80.6 & [76.9, 84.2] & 9.41 & -- & 94.5 & 84.3 & 76.4 \\
  \cmidrule(l){2-10}
  \rowcolor{oursrow}  & POPE & \textbf{Trace reliability (ours)} & \textbf{80.5} & [77.9, 83.0] & \textbf{3.11} & \textbf{7.9} & \textbf{97.7} & \textbf{96.5} & \textbf{93.7} \\
   &  & \scout single-view conf. & 56.4 & [53.0, 59.7] & 8.38 & 9.2 & 92.0 & 92.4 & 91.4 \\
   &  & SpecEyes separability & 44.4 & [42.1, 46.6] & 10.64 & -- & 88.6 & 87.5 & 89.1 \\
  \midrule
  \rowcolor{oursrow} Thyme-RL & V* & \textbf{Trace reliability (ours)} & \textbf{93.7} & [89.5, 96.9] & \textbf{1.10} & \textbf{4.1} & \textbf{100.0} & \textbf{100.0} & \textbf{95.3} \\
   &  & \scout single-view conf. & 83.3 & [76.0, 89.9] & 2.24 & 16.4 & 99.0 & 96.3 & 91.9 \\
   &  & SpecEyes separability & 77.6 & [71.5, 83.4] & 5.28 & -- & 96.9 & 90.3 & 85.5 \\
  \cmidrule(l){2-10}
  \rowcolor{oursrow}  & HR & \textbf{Trace reliability (ours)} & \textbf{88.6} & [85.4, 91.7] & \textbf{5.89} & \textbf{11.2} & \textbf{98.6} & \textbf{91.2} & \textbf{81.4} \\
   &  & \scout single-view conf. & 77.8 & [72.7, 82.5] & 8.76 & 24.8 & 94.3 & 84.9 & 77.7 \\
   &  & SpecEyes separability & 70.3 & [66.9, 73.6] & 10.46 & -- & 90.3 & 81.4 & 78.8 \\
  \cmidrule(l){2-10}
  \rowcolor{oursrow}  & POPE & \textbf{Trace reliability (ours)} & \textbf{80.4} & [77.9, 82.8] & \textbf{3.32} & \textbf{7.9} & \textbf{97.4} & \textbf{96.5} & \textbf{93.5} \\
   &  & \scout single-view conf. & 57.0 & [53.5, 60.5] & 8.46 & 9.2 & 91.9 & 92.2 & 91.3 \\
   &  & SpecEyes separability & 48.2 & [45.9, 50.6] & 10.04 & -- & 88.6 & 89.7 & 89.9 \\
  \bottomrule
  \end{tabular}
\end{table*}

\begin{table*}[t]
  \begin{minipage}[t]{0.485\linewidth}
  \centering
  \caption{Change in AUROC (points) without one part of the trace score, averaged over the three benchmarks, at batch size 1.}
  \label{tab:suppl_trust_ablation}
  \footnotesize
  \setlength{\tabcolsep}{2.8pt}
  \begin{tabular}{l cc}
  \toprule
  Variant & DeepEyes-7B & Thyme-RL \\
  \midrule
  w/o certified-first order & $-$0.7 & $-$0.8 \\
  w/o normalized pool & $-$2.0 & $-$0.1 \\
  w/o \overseer view & $-$0.5 & $-$0.5 \\
  w/o \scout views & $-$9.6 & $-$8.4 \\
  w/o correction view & $-$1.5 & $-$0.4 \\
  \bottomrule
  \end{tabular}
  \end{minipage}\hfill
  \begin{minipage}[t]{0.485\linewidth}
  \centering
  \caption{AUROC (\%) of reliability scores for detecting wrong answers of \methodfast at batch size 1. Two-view conf.: \scout confidence averaged over the whole image and the glimpse.}
  \label{tab:suppl_trust_scores}
  \footnotesize
  \setlength{\tabcolsep}{2.8pt}
  \begin{tabular}{l ccc ccc}
  \toprule
  AUROC $\uparrow$ & \multicolumn{3}{c}{DeepEyes-7B} & \multicolumn{3}{c}{Thyme-RL} \\
  \cmidrule(lr){2-4} \cmidrule(l){5-7}
  Score & V* & HR & POPE & V* & HR & POPE \\
  \midrule
  SpecEyes separability & 77.5 & \second{80.6} & 44.4 & 77.6 & 70.3 & 48.2 \\
  \scout single-view conf. & 84.8 & 77.1 & 56.4 & 83.3 & 77.8 & 57.0 \\
  \scout two-view conf. & \second{89.5} & 79.9 & \second{68.1} & \second{89.3} & \second{81.6} & \second{69.0} \\
  \rowcolor{oursrow} \textbf{Trace reliability (ours)} & \textbf{93.4} & \textbf{87.1} & \textbf{80.5} & \textbf{93.7} & \textbf{88.6} & \textbf{80.4} \\
  \midrule
  \rowcolor{grouprow} \multicolumn{7}{l}{\textit{Questions sent to the \overseer}} \\
  \overseer confidence & 38.1 & 68.3 & 75.7 & 54.4 & 65.2 & 55.5 \\
  \rowcolor{oursrow} \textbf{Trace reliability (ours)} & \textbf{85.6} & \textbf{76.9} & \textbf{84.1} & \textbf{85.8} & \textbf{79.0} & \textbf{84.9} \\
  \bottomrule
  \end{tabular}
  \end{minipage}
\end{table*}

\begin{table*}[t]
  \centering
  \caption{The DeepEyes-7B agent loop on the questions with the lowest trace score. Agent loop: \% of all questions.}
  \label{tab:suppl_agent_fallback}
  \small
  \setlength{\tabcolsep}{6pt}
  \begin{tabular}{l c cccccc c c}
  \toprule
   & Agent & & \multicolumn{2}{c}{HR-Bench} & \multicolumn{3}{c}{POPE} & & \\
  \cmidrule(lr){4-5} \cmidrule(lr){6-8}
  Method & loop & V* & 4K & 8K & Rand. & Pop. & Adv. & Mean & Spd. $\uparrow$ \\
  \midrule
  \rowcolor{grouprow} \multicolumn{10}{l}{\textit{DeepEyes-7B as the expert}} \\
  \rowcolor{oursrow} \methodfast & 0\% & 91.6 & 79.9 & 75.8 & 92.8 & 90.4 & 88.8 & 86.5 & \spd{100}{9.9$\times$} \\
  \rowcolor{oursrow} \method & 0\% & 92.7 & 81.0 & 76.8 & 93.1 & 90.6 & 89.0 & 87.2 & \spd{75}{7.7$\times$} \\
  \methodfast + agent loop & 1\% & 91.6 & 79.6 & 75.5 & 92.8 & 90.3 & 88.5 & 86.4 & \spd{90}{8.9$\times$} \\
  \methodfast + agent loop & 5\% & 92.7 & 78.4 & 75.3 & 93.0 & 90.0 & 88.2 & 86.2 & \spd{55}{5.8$\times$} \\
  \methodfast + agent loop & 10\% & 92.1 & 76.9 & 73.0 & 93.4 & 90.1 & 88.0 & 85.6 & \spd{35}{4.2$\times$} \\
  \methodfast + agent loop & 20\% & 90.6 & 74.5 & 71.3 & 93.6 & 88.8 & 86.1 & 84.1 & \spd{20}{2.9$\times$} \\
  DeepEyes-7B agent & 100\% & 89.0 & 74.6 & 70.8 & 92.1 & 85.5 & 81.8 & 82.3 & 1.0$\times$ \\
  \bottomrule
  \end{tabular}
\end{table*}

\textbf{Relation to existing techniques.}
\method changes the common form of four techniques, routing between a small and a large model, visual cropping, a product of experts, and an early exit.
The ablations isolate each change.
\begin{itemize}
  \setlength{\itemsep}{0pt}
  \item \textit{Routing and speculation.} A cascade or answer speculation routes on the confidence of the small model's answer to the same input. \method speculates the next observation and routes on whether the answer is unchanged under it. With the confidence rule on the same \scout outputs, 700 wrong answers are returned with no \overseer call instead of 240, and V* accuracy falls with both experts (Table 2 of the main paper).
  \item \textit{Visual cropping.} Cropping methods choose a crop and answer from it. In \method the crop is a draft that a second model checks and corrects, and only one of the two models answers from it. A single model that drafts and reads its own glimpse is less accurate than \methodfast on every split (\cref{tab:suppl_zoom_refine}), and removing the check and correction lowers V* accuracy with both experts.
  \item \textit{Product of experts.} A model ensemble usually gives every member the same input. \method assigns the views so that the errors of the two models are less correlated. Sharing the glimpse raises the error correlation on all three benchmarks (\cref{tab:suppl_view_assignment}), and the fused answer is then correct on 33 fewer questions although the \overseer alone is correct on 204 more.
  \item \textit{Early exit.} An early exit thresholds one confidence. The certified exit adds a test on new evidence. Its accepted errors are a subset of those of the confidence rule at the same threshold (\cref{prop:suppl_exit_subset}), and none of its 681 answers is wrong.
\end{itemize}
The parts also share one trace.
The two \scout answers that route a question are views of the fusion and terms of the reliability score, so the exit test and the score add no model call to those that answer the question.

\section{Reliability}
\label{sec:suppl_reliability}

\textbf{Risk and coverage.}
\cref{fig:suppl_risk_coverage,tab:suppl_trust} extend Figure 9 of the main paper to both experts.
AURC is the area under the risk-coverage curve.
The certified exits take the top of the ranking with no wrong answer, 96 on V*, 573 on HR-Bench, and 12 on POPE.
For both experts and on all three benchmarks, the trace score is the best score on every metric.
At 50\% coverage, its selective accuracy is 100.0\% on V* with both experts, and 98.3\% with DeepEyes-7B and 98.6\% with Thyme-RL on HR-Bench.

\textbf{Confidence intervals.}
Each AUROC in \cref{tab:suppl_trust} has a 95\% percentile interval from 2,000 bootstrap resamples of the images~\cite{efron1979bootstrap}, the unit of \cref{sec:suppl_zero_errors}.
On the same resamples, the AUROC gain of the trace score over the SpecEyes score and over the single-view confidence of the \scout has a 95\% interval above zero in all twelve comparisons, for example $[8.9, 23.4]$ points over the SpecEyes score on the 191 V* questions with DeepEyes-7B.

\textbf{Calibration.}
The trace score is a ranking of the answers, and AUROC, AURC, and selective accuracy depend only on its order.
Read as a probability, with the score of a certified exit set to one, it is also better calibrated than the single-view confidence of the \scout. Its expected calibration error~\cite{guo2017calibration} over ten equal-width bins is lower on every benchmark with both experts, 3.5\% against 15.8\% on V* with DeepEyes-7B.

\begin{table*}[t]
  \centering
  \caption{\method on multiple-choice questions without parts of its search. Mean: mean accuracy over the three splits. Spd.: speedup over the agent on the three splits together.}
  \label{tab:suppl_agentic_ablation}
  \small
  \setlength{\tabcolsep}{4pt}
  \begin{tabular}{l ccccc ccccc}
  \toprule
   & \multicolumn{5}{c}{DeepEyes-7B} & \multicolumn{5}{c}{Thyme-RL} \\
  \cmidrule(lr){2-6} \cmidrule(l){7-11}
  Variant & V* & 4K & 8K & Mean & Spd. $\uparrow$ & V* & 4K & 8K & Mean & Spd. $\uparrow$ \\
  \midrule
  \methodfast (Table 1) & \second{91.6} & 79.9 & 75.8 & 82.4 & \spd{60}{6.2$\times$} & \second{89.5} & 78.6 & 74.4 & 80.8 & \spd{40}{4.5$\times$} \\
  \rowcolor{oursrow} \textbf{\method} & \textbf{92.7} & \textbf{81.0} & \second{76.8} & \textbf{83.5} & \spd{35}{4.0$\times$} & \textbf{90.1} & \second{80.0} & \second{75.4} & \second{81.8} & \spd{20}{2.8$\times$} \\
  w/o agent answer & \textbf{92.7} & 79.8 & \second{76.8} & \second{83.1} & \spd{35}{4.2$\times$} & 89.0 & \second{80.0} & 75.1 & 81.4 & \spd{20}{3.0$\times$} \\
  w/o grounding & \second{91.6} & \textbf{81.0} & 76.4 & 83.0 & \spd{40}{4.5$\times$} & 89.0 & 79.9 & 74.4 & 81.1 & \spd{20}{3.0$\times$} \\
  w/o grounding and agent answer & 90.6 & \second{80.3} & 76.3 & 82.4 & \spd{40}{4.8$\times$} & 88.0 & 79.8 & 74.5 & 80.7 & \spd{25}{3.3$\times$} \\
  \midrule
  \methodfast + agent loop on searched & \textbf{92.7} & 78.0 & \textbf{76.9} & 82.5 & \spd{15}{2.2$\times$} & \second{89.5} & \textbf{80.1} & \textbf{76.9} & \textbf{82.2} & \spd{10}{1.9$\times$} \\
  \bottomrule
  \end{tabular}
\end{table*}

\begin{table*}[t]
  \centering
  \caption{\method on POPE without terms of its yes/no rule. Mean and Spd. as in \cref{tab:suppl_agentic_ablation}.}
  \label{tab:suppl_agentic_pope}
  \small
  \setlength{\tabcolsep}{4pt}
  \begin{tabular}{l ccccc ccccc}
  \toprule
   & \multicolumn{5}{c}{DeepEyes-7B} & \multicolumn{5}{c}{Thyme-RL} \\
  \cmidrule(lr){2-6} \cmidrule(l){7-11}
  Variant & Rand. & Pop. & Adv. & Mean & Spd. $\uparrow$ & Rand. & Pop. & Adv. & Mean & Spd. $\uparrow$ \\
  \midrule
  \methodfast (Table 1) & 92.8 & \second{90.4} & 88.8 & \second{90.7} & \spd{100}{13.2$\times$} & 92.6 & 89.8 & 88.3 & \second{90.3} & \spd{80}{8.3$\times$} \\
  \rowcolor{oursrow} \textbf{\method} & 93.1 & \textbf{90.6} & \second{89.0} & \textbf{90.9} & \spd{100}{12.6$\times$} & 92.8 & \textbf{90.1} & \second{88.6} & \textbf{90.5} & \spd{75}{7.9$\times$} \\
  w/o \scout union & 91.8 & 90.2 & \textbf{89.2} & 90.4 & \spd{100}{12.6$\times$} & 91.4 & 89.7 & \textbf{88.8} & 90.0 & \spd{75}{7.9$\times$} \\
  w/o \overseer whole image & \second{93.3} & 90.2 & 88.1 & 90.5 & \spd{100}{12.6$\times$} & \textbf{93.1} & \second{89.9} & 87.9 & \second{90.3} & \spd{75}{7.9$\times$} \\
  \midrule
  \methodfast + agent loop on searched & \textbf{93.5} & 89.4 & 86.8 & 89.9 & \spd{50}{5.7$\times$} & \second{93.0} & 89.6 & 87.1 & 89.9 & \spd{40}{4.6$\times$} \\
  \bottomrule
  \end{tabular}
\end{table*}

\begin{table*}[t]
  \centering
  \caption{AURC of the answers of \methodfast and \method in their Table 1 runs.}
  \label{tab:suppl_agentic_trust}
  \small
  \setlength{\tabcolsep}{6pt}
  \begin{tabular}{l ccc ccc}
  \toprule
  AURC (\%) $\downarrow$ & \multicolumn{3}{c}{DeepEyes-7B} & \multicolumn{3}{c}{Thyme-RL} \\
  \cmidrule(lr){2-4} \cmidrule(l){5-7}
  Method & V* & HR & POPE & V* & HR & POPE \\
  \midrule
  \methodfast & 1.22 & 5.55 & 3.12 & 1.47 & 5.87 & 3.34 \\
  \rowcolor{oursrow} \textbf{\method} & \textbf{1.21} & \textbf{4.96} & \textbf{3.11} & 1.47 & \textbf{5.29} & \textbf{3.33} \\
  \bottomrule
  \end{tabular}
\end{table*}

\begin{table*}[t]
  \centering
  \caption{Accuracy (\%) on the questions that \method searches. Fixed / broken: answers that the search turns correct (+) and wrong ($-$).}
  \label{tab:suppl_search_questions}
  \small
  \setlength{\tabcolsep}{6pt}
  \begin{tabular}{l cccc cccc}
  \toprule
   & \multicolumn{4}{c}{DeepEyes-7B} & \multicolumn{4}{c}{Thyme-RL} \\
  \cmidrule(lr){2-5} \cmidrule(l){6-9}
   & V* & HR & POPE & All & V* & HR & POPE & All \\
  \midrule
  Searched questions & 40 & 478 & 908 & 1{,}426 & 44 & 500 & 921 & 1{,}465 \\
  \midrule
  \methodfast & 70.0 & 51.9 & 64.0 & 60.1 & 65.9 & 50.6 & 60.6 & 57.3 \\
  Agent loop & \textbf{75.0} & 50.6 & 56.4 & 55.0 & 65.9 & \textbf{57.0} & 57.3 & 57.5 \\
  \rowcolor{oursrow} \textbf{\method} & \textbf{75.0} & \textbf{55.4} & \textbf{66.4} & \textbf{63.0} & \textbf{68.2} & 54.4 & \textbf{63.1} & \textbf{60.3} \\
  \midrule
  Fixed / broken & +3/$-$1 & +36/$-$19 & +59/$-$37 & +98/$-$57 & +3/$-$2 & +27/$-$8 & +62/$-$39 & +92/$-$49 \\
  \bottomrule
  \end{tabular}
\end{table*}

\begin{table*}[t]
  \centering
  \caption{\method with a stronger agent on its searched multiple-choice questions. Mean: mean accuracy over the six splits. Spd.: speedup over the agent of the expert.}
  \label{tab:suppl_strong_agent}
  \small
  \setlength{\tabcolsep}{6pt}
  \begin{tabular}{l cccccc c c}
  \toprule
   & & \multicolumn{2}{c}{HR-Bench} & \multicolumn{3}{c}{POPE} & & \\
  \cmidrule(lr){3-4} \cmidrule(lr){5-7}
  Method & V* & 4K & 8K & Rand. & Pop. & Adv. & Mean & Spd. $\uparrow$ \\
  \midrule
  \rowcolor{grouprow} \multicolumn{9}{l}{\textit{DeepEyes-7B as the expert}} \\
  \methodfast & 91.6 & 79.9 & 75.8 & 92.8 & 90.4 & 88.8 & 86.5 & \spd{100}{9.9$\times$} \\
  \rowcolor{oursrow} \textbf{\method} & \textbf{92.7} & 81.0 & 76.8 & \textbf{93.1} & \textbf{90.6} & \textbf{89.0} & 87.2 & \spd{75}{7.7$\times$} \\
  \method + SenseNova-MARS-8B on searched & \textbf{92.7} & \textbf{84.0} & \textbf{80.1} & \textbf{93.1} & \textbf{90.6} & \textbf{89.0} & \textbf{88.2} & \spd{40}{4.7$\times$} \\
  \midrule
  \rowcolor{grouprow} \multicolumn{9}{l}{\textit{Thyme-RL as the expert}} \\
  \methodfast & 89.5 & 78.6 & 74.4 & 92.6 & 89.8 & 88.3 & 85.5 & \spd{60}{6.3$\times$} \\
  \rowcolor{oursrow} \textbf{\method} & 90.1 & 80.0 & 75.4 & \textbf{92.8} & \textbf{90.1} & \textbf{88.6} & 86.2 & \spd{40}{4.7$\times$} \\
  \method + SenseNova-MARS-8B on searched & \textbf{92.1} & \textbf{83.1} & \textbf{78.9} & \textbf{92.8} & \textbf{90.1} & \textbf{88.6} & \textbf{87.6} & \spd{20}{2.7$\times$} \\
  \bottomrule
  \end{tabular}
\end{table*}

\textbf{Parts of the trace score.}
\cref{tab:suppl_trust_ablation} removes one part of the trace score at a time, scores the same answers, and averages the change in AUROC over the three benchmarks.
Removing any part lowers the AUROC with both experts.
The \scout views matter most. Without them, the AUROC falls by 9.6 points with DeepEyes-7B and 8.4 points with Thyme-RL, and by 18.6 and 13.3 points on V*.

\textbf{Other scores.}
\cref{tab:suppl_trust_scores} compares the trace score with the acceptance score of SpecEyes, which its agent-loop answers lack, and with the confidence of the \scout on one view or averaged over its two views.
The two \scout confidences exist for every answer of \methodfast, so these rows compare scores on the same answers at full coverage.
The trace score has the highest AUROC in every column.
On the questions sent to the \overseer, the confidence of the \overseer itself has an AUROC of 38.1\% on V* with DeepEyes-7B and 54.4\% with Thyme-RL, against 85.6\% and 85.8\% for the trace score.

\textbf{POPE.}
On POPE, the trace score raises the AUROC from 44.4\% and 48.2\% for the SpecEyes score to 80.5\% and 80.4\%.
A POPE question asks whether one object is present, and an absent object has no region to zoom into.
As a result, 43.3\% of the POPE questions end at a single-view exit, where the \scout drafts no glimpse, the trace holds one view, and the score is the probability that the \scout gives to its answer.
Of the 234 wrong answers on this route, 225 answer that an object labeled as present is absent, and the corrected labels of RePOPE~\cite{neuhaus2025repope} change the label of 137 of them (\cref{sec:suppl_repope}).
On the POPE questions sent to the \overseer, where the trace holds the views of both models, the AUROC of the trace score is 84.1\% and 84.9\%, against 85.6\% and 85.8\% on V* and 76.9\% and 79.0\% on HR-Bench (\cref{tab:suppl_trust_scores}).

\textbf{Evidence region.}
V* annotates the target objects of every question, which tests the evidence region $e$ of \methodfast at batch size 1.
The region is a crop on 188 of the 191 questions, and it contains the center of every annotated target on 149 of them (79.3\%) with DeepEyes-7B and on 140 (74.5\%) with Thyme-RL.
The zoom crops of the DeepEyes-7B agent contain the targets on 123 of the 191 questions (64.4\%).
The answer also follows the region. It is correct on 96.6\% and 97.1\% of the questions whose region contains the targets, against 74.4\% and 70.8\% of the others, and a crop moved away from the target lowers the accuracy of the \scout (Figure 7 of the main paper).
The check and correction improve the region. Of the 34 glimpses that DeepEyes-7B corrects, 10 contain the targets before the correction and 22 after it, and of the 19 that Thyme-RL corrects, 6 before and 9 after.

\textbf{Agent loop on low-score questions.}
In \cref{tab:suppl_agent_fallback}, the DeepEyes-7B agent answers the questions with the lowest trace score, and \methodfast answers the rest.
The runs use released batch sizes.
``Mean'' averages accuracy over the six splits, and the speedup is over the agent, at its mean time per question on each split.
\method has the highest mean accuracy, 87.2\%, at 7.7$\times$, and \methodfast alone has the largest speedup, 9.9$\times$, with the next highest mean accuracy, 86.5\%.
Every mix has a higher mean accuracy than the agent alone.
Handing low-score questions to the agent loop therefore lowers the mean accuracy, and \method instead searches the flagged questions without the loop (\cref{sec:suppl_agent}).

\section{Analysis of the Evidence Search}
\label{sec:suppl_agent}

This section analyzes the bounded evidence search of \method, which \methodfast skips, on its runs in Table 1 of the main paper.
\cref{sec:suppl_impl} gives its implementation.

\begin{figure*}[t]
  \centering
  \includegraphics[width=\linewidth]{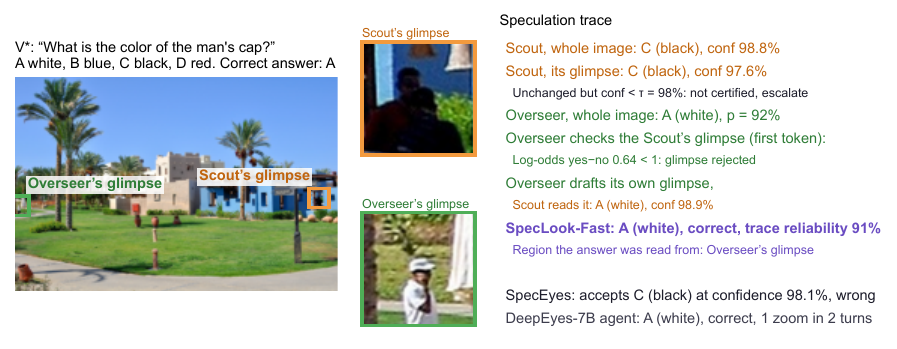}
  \caption{A \methodfast trace on V* with DeepEyes-7B at batch size 1. SpecEyes resizes the image by its own rule, so its confidence differs from that of the \scout.}
  \label{fig:suppl_qualitative}
\end{figure*}

\begin{figure*}[!t]
  \centering
  \includegraphics[width=0.94\linewidth]{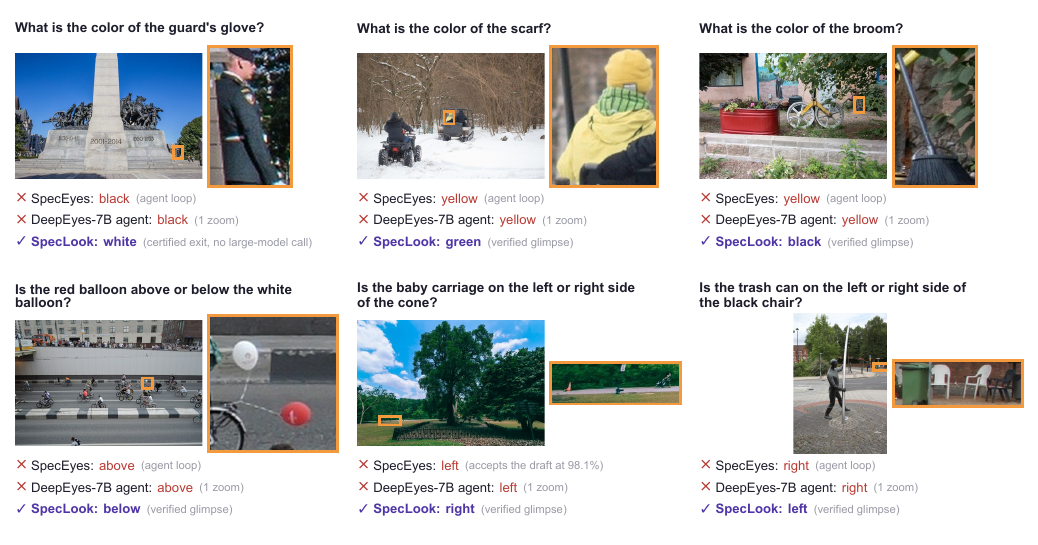}
  \caption{V* questions that \method answers correctly and that SpecEyes and the DeepEyes-7B agent answer wrongly. Orange: the glimpse of the \scout, enlarged on the right.}
  \label{fig:suppl_qual_vstar}
\end{figure*}

\textbf{Parts of the search.}
In \cref{tab:suppl_agentic_ablation}, removing the grounding and the agent answer together lowers the accuracy on V* by 2.1 points with both experts. The mean accuracy then falls from 83.5\% to 82.4\% with DeepEyes-7B and from 81.8\% to 80.7\% with Thyme-RL, against 82.4\% and 80.8\% for \methodfast.
Removing either part alone lowers the accuracy in 8 of the 12 cells and raises it in none.
The agent loop on the searched questions is timed at the mean time per question of the agent on each split.
Handing the searched questions to the agent loop instead lowers the speedup from 4.0$\times$ to 2.2$\times$ with DeepEyes-7B and from 2.8$\times$ to 1.9$\times$ with Thyme-RL.
With DeepEyes-7B, it is also less accurate than \method, 82.5\% against 83.5\% in mean accuracy.

\textbf{Terms of the yes/no rule.}
In \cref{tab:suppl_agentic_pope}, the full rule has the highest mean accuracy with both experts, 90.9\% with DeepEyes-7B and 90.5\% with Thyme-RL.
Without the union of the \scout views, accuracy on the random split falls by 1.3 points with DeepEyes-7B and 1.4 points with Thyme-RL, and without the whole-image answer of the \overseer, accuracy on the adversarial split falls by 0.9 and 0.7 points.
Handing the searched questions to the agent loop lowers the accuracy on the adversarial split from 89.0\% to 86.8\% with DeepEyes-7B and from 88.6\% to 87.1\% with Thyme-RL, and the speedup from 12.6$\times$ to 5.7$\times$ and from 7.9$\times$ to 4.6$\times$.

\textbf{Reliability.}
\method scores a searched answer with the minimum of its trace score and the normalized pool of the fused views with equal weights, or the sigmoid of the log-odds of the yes/no rule, and keeps the scores of the other answers.
In \cref{tab:suppl_agentic_trust}, its AURC is at most that of \methodfast in every column.
On HR-Bench, the AURC falls from 5.55\% to 4.96\% with DeepEyes-7B and from 5.87\% to 5.29\% with Thyme-RL.
Certified exits are never searched, so they keep their answers and scores.

\textbf{Searched questions.}
\cref{tab:suppl_search_questions} reports the questions that \method searches, 13.2\% of all questions with DeepEyes-7B and 13.6\% with Thyme-RL.
The search changes no other answer, and the other answers are 93.1\% correct with both experts.
On the searched questions, the search raises the accuracy from 60.1\% to 63.0\% with DeepEyes-7B and from 57.3\% to 60.3\% with Thyme-RL.
The search improves more images than it worsens with both experts, 49 against 30 and 48 against 29.
These questions are hard for the agent as well. Its loop is correct on 55.0\% and 57.5\% of them, below \method with both experts.

\textbf{Stronger agent.}
The search can call an agent other than the expert.
In \cref{tab:suppl_strong_agent}, SenseNova-MARS-8B, an agentic MLLM built on Qwen3-VL-8B, answers the searched multiple-choice questions with its released tool loop from the whole image at batch size 1.
The loop follows the released prompt with greedy decoding, at most 10 turns, and at most 1,024 new tokens per turn. Its zoom tool is executed, and its two web search tools return no result.
Its answer replaces the answer of \method, which is kept when the loop ends without a valid option.
Yes/no questions make no agent call, so the POPE columns are unchanged.
The time is the Table 1 time of \method plus the time of every loop, measured in one end-to-end run per split.
HR-Bench accuracy rises by 3.0 and 3.3 points with DeepEyes-7B and by 3.1 and 3.5 points with Thyme-RL.
The mean accuracy reaches 88.2\% and 87.6\%, which is 1.7 and 2.1 points above \methodfast, at 4.7$\times$ and 2.7$\times$ the speed of the agents.
The search thus gains from a stronger agent with no change to its setting, and the accuracy of \method over \methodfast grows with the agent that the search calls.

\section{Qualitative Examples}
\label{sec:suppl_qualitative}

\begin{figure*}[!t]
  \centering
  \includegraphics[width=0.94\linewidth]{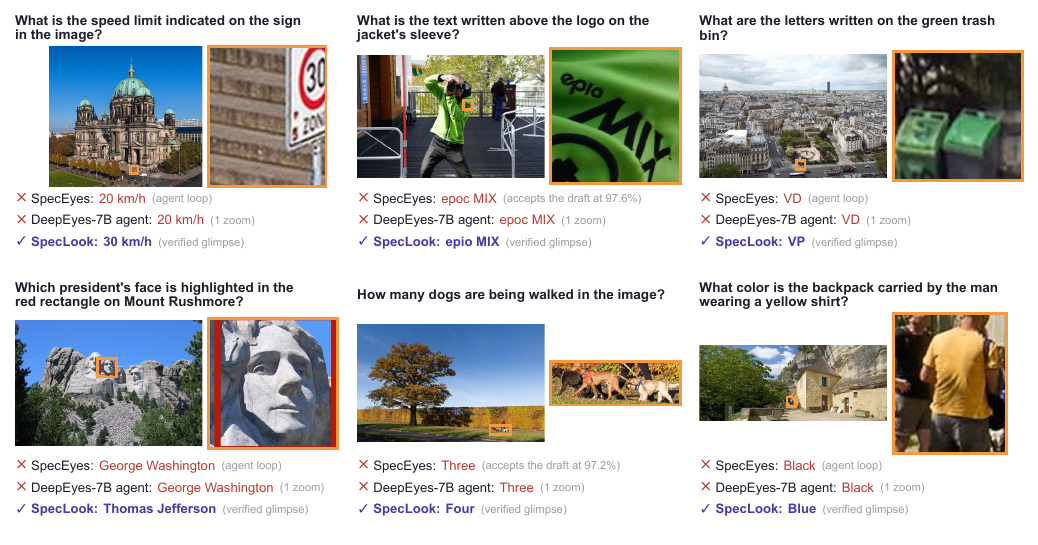}
  \caption{HR-Bench questions, shown as in \cref{fig:suppl_qual_vstar}.}
  \label{fig:suppl_qual_hr}
\end{figure*}

\begin{figure*}[t]
  \centering
  \includegraphics[width=0.94\linewidth]{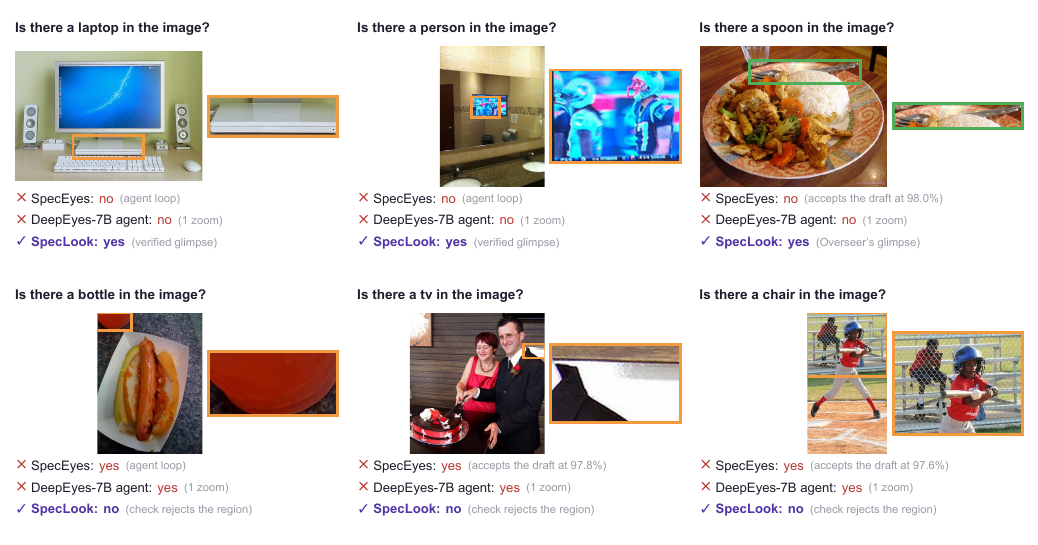}
  \caption{POPE questions, shown as in \cref{fig:suppl_qual_vstar}. Top: present objects. Bottom: absent objects, where the check rejects the proposed region. Green: the glimpse of the \overseer.}
  \label{fig:suppl_qual_pope}
\end{figure*}

\begin{figure*}[t]
  \centering
  \includegraphics[width=0.94\linewidth]{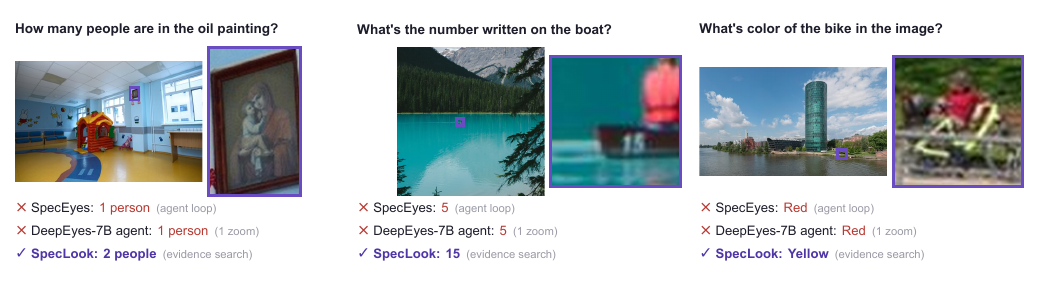}
  \caption{Questions answered by the evidence search of \method. Violet: the region grounded by the search.}
  \label{fig:suppl_qual_search}
\end{figure*}

\textbf{Speculation trace.}
\cref{fig:suppl_qualitative} shows the trace of one V* question with DeepEyes-7B at batch size 1.
The \scout answers black from the whole image and from its glimpse, with confidences of 98.8\% and 97.6\%.
The answer is unchanged after zooming but below $\tau$, so the question is escalated.
The \overseer answers white from the whole image, and its check rejects the glimpse of the \scout, whose log-odds of 0.64 is below $\delta = 1$.
The \overseer then drafts its own glimpse, from which the \scout answers white with a confidence of 98.9\%.
The fused answer is white, which is correct, with a trace reliability of 91\%, and the region it was read from is the glimpse of the \overseer.
\method searches this question and keeps the correct answer white.
SpecEyes, which resizes the image by its own rule before the same 2B model reads it, accepts the wrong answer black at a confidence of 98.1\% instead of the 98.8\% of the \scout, and the DeepEyes-7B agent answers correctly after one zoom in two turns.

\textbf{Comparison on single questions.}
\cref{fig:suppl_qual_vstar,fig:suppl_qual_hr,fig:suppl_qual_pope} show questions from the runs of Table 1 of the main paper with DeepEyes-7B that \method answers correctly and that both SpecEyes and the DeepEyes-7B agent answer wrongly.
Each panel gives the whole image, the region that the answer of \method was read from, and the answer of each method.
On V* and HR-Bench, the target covers a small part of the image, such as a glove, a logo on a sleeve, or a trash bin in an aerial view of a city.
The agent zooms once and still returns a wrong option, and SpecEyes either accepts a wrong draft with high confidence or falls back to the same agent loop.
\method reads the verified glimpse as a closer view and returns the correct option, in one case at a certified exit with no large-model call.
On POPE, the top row shows present objects that the other two methods miss, and the bottom row shows absent objects that they report.
For the absent objects, the check of the \overseer rejects the region that the \scout proposed, and \method answers no.
\cref{fig:suppl_qual_search} shows three questions on which the evidence search of \method grounds the target, and \method then answers correctly.

\section{Formal and Statistical Analysis}
\label{sec:suppl_analysis}

This section derives the properties of routing, fusion, the reliability score, and cost behind Section 3 of the main paper, proves the proposition stated there, and tests the main results statistically.

\subsection{Setting}
\label{sec:suppl_scope}

Let $Y\in\mathcal A$ be the correct answer.
The models, prompts, decoding rules, and thresholds are fixed, and probabilities refer to a population of image-question pairs.
The question is suppressed in $p_m(a\mid v)$, as in the main paper.
The \scout first returns $(a_0,c_0)$ on $I$.
Let $\Gamma$ denote the event that its draft gives a valid, nonempty glimpse $g$.
On $\Gamma$, its closer view $(I,I[g])$ returns $(a_1,c_1)$.
A recovered glimpse enters only on escalated questions, so it never creates a certified exit.
The two answer views share the whole image and differ in the glimpse, which enters only the answer of the \scout.
The \overseer reads the crop in the glimpse check, whose verdict decides whether a corrective view is added.

\subsection{Accepted Errors and Selective Risk}
\label{sec:suppl_exit_analysis}

Define the confidence-only acceptance event $C_\tau=\{c_0\geq\tau\}$ and the certified event
\begin{equation}
  D_\tau=\Gamma\cap\{a_0=a_1\}\cap\{\min(c_0,c_1)\geq\tau\}.
  \label{eq:suppl_cert_event}
\end{equation}
All early exits form $H_\tau=D_\tau\cup(C_\tau\cap\Gamma^c)$, with answer $a_0$ on either branch.

\begin{proposition}[Accepted-error inclusion]
\label{prop:suppl_exit_subset}
For fixed \scout outputs and the same threshold $\tau$, the wrong early exits of \method are a subset of the wrong exits of the confidence-only rule:
\begin{equation}
  H_\tau\cap\{a_0\ne Y\}\ \subseteq\ C_\tau\cap\{a_0\ne Y\}.
\end{equation}
The inclusion holds on every evaluation set, with no calibration or independence assumption.
\end{proposition}

\begin{proof}
Both branches of $H_\tau$ require $c_0\geq\tau$ and return $a_0$.
Hence $H_\tau\subseteq C_\tau$, and intersecting both sides with the same error event proves the claim.
\end{proof}

The inclusion bounds the number of accepted errors.
The error rate among accepted answers also depends on how many correct drafts the test keeps.
To state this rate, restrict attention to $B=C_\tau\cap\Gamma$ with $\Pr(B)>0$.
Let $F=\{a_0=a_1,c_1\geq\tau\}$ be the closer-view test, so that $D_\tau=B\cap F$, and define
\begin{align}
  \rho&=\Pr(a_0\ne Y\mid B),\nonumber\\
  \alpha&=\Pr(F\mid a_0\ne Y,B),\nonumber\\
  \beta&=\Pr(F\mid a_0=Y,B).
\end{align}
Assume $0<\rho<1$ and $\rho\alpha+(1-\rho)\beta>0$.
Partitioning the accepted events by correctness gives
\begin{equation}
  \Pr(a_0\ne Y\mid D_\tau)
  =\frac{\rho\alpha}{\rho\alpha+(1-\rho)\beta}.
  \label{eq:suppl_selective_risk}
\end{equation}
The right side is at most $\rho$ exactly when $\alpha\leq\beta$.
Multiplying by the positive denominator and canceling $\rho(1-\rho)$ reduces the inequality to this condition.
The closer view thus lowers the error rate of accepted drafts exactly when it keeps wrong drafts less often than correct ones.
Figure 3 of the main paper measures this directly. Every confident wrong draft with a glimpse fails the test, so $\alpha=0$ on the evaluated questions, while the certified exits keep 96 drafts on V*, 573 on HR-Bench, and 12 on POPE.

\subsection{What the Product of Experts Computes}
\label{sec:suppl_pool_analysis}

On an escalated question, let $\mathcal V$ contain the \scout on its closer view, or on $I$ if no closer view is available, the \overseer on $I$, and the \scout on a nonempty corrective view when one is drafted.
A recovered glimpse takes the place of the initial glimpse in this set.
Write $n=|\mathcal V|$ and $p_i(a)=p_m(a\mid v)$ for its members, and assume $p_i(a)>0$ for every $a\in\mathcal A$.
The multiple-choice rule maximizes $\sum_i\log p_i(a)$.
For any candidate $y$ and alternative $b$, their score difference is
\begin{equation}
  M(y,b)=\sum_{i=1}^{n}\log\frac{p_i(y)}{p_i(b)}.
  \label{eq:suppl_fusion_margin}
\end{equation}
The answer $y$ is selected when $M(y,b)\geq0$ for every $b$, with the decoding tie rule deciding ties.
Each view adds its own log ratio to every margin, with no independence assumption.
A corrective view that favors the right answer can therefore reverse a wrong fused answer, and its weight equals that of every other view.

\textbf{A conditional Bayesian interpretation.}
Fix a routing event that determines $n$ and the types of view.
Suppose that the observations $X_i$ are conditionally independent given $Y$ on this stratum, and that each $p_i$ is the posterior $\Pr(Y=a\mid X_i)$ with prior $\omega(a)>0$.
Bayes' rule then gives
\begin{equation}
  \Pr(Y=a\mid X_1,\dots,X_n)
  \propto \omega(a)^{1-n}\prod_{i=1}^{n}p_i(a).
  \label{eq:suppl_bayes_pool}
\end{equation}
To verify this expression, write $p_i(a)\propto\omega(a)f_i(X_i\mid a)$.
The factor $\omega(a)^{1-n}$ removes the extra copies of the prior and leaves the joint posterior.
With a uniform prior, maximizing the product is the Bayes decision.
The views share the whole image, which is why \method assigns the glimpse to one model only.
The fused answer is also more accurate than either model alone. On the 1,095 escalated multiple-choice questions at batch size 1, it is correct on 67.8\% with DeepEyes-7B and 64.3\% with Thyme-RL, against 60.3\% for the \scout on its view and 53.9\% and 53.0\% for the \overseer on the whole image.

\subsection{Joint Errors and View Assignment}
\label{sec:suppl_complementarity}

The following result restates and proves the proposition of the main paper.

\begin{proposition}[Oracle bound for selecting between two answers]
\label{prop:suppl_complementarity}
On a fixed population, let $E_S,E_L$ be the error indicators of two answers, with $0<e_S=\mathbb E[E_S]<1$ and $0<e_L=\mathbb E[E_L]<1$.
Set $\sigma_S=\sqrt{e_S(1-e_S)}$, $\sigma_L=\sqrt{e_L(1-e_L)}$, and $\phi=\operatorname{Corr}(E_S,E_L)$.
Any rule that returns one of these answers has accuracy at most
\begin{equation}
  1-e_Se_L-\phi\sigma_S\sigma_L.
  \label{eq:suppl_oracle_ceiling}
\end{equation}
Keep the first answer fixed and replace the second by one with $0<e_{L'}<e_L$, standard deviation $\sigma_{L'}$, and correlation $\phi'$ with $E_S$.
The replacement lowers this bound exactly when
\begin{equation}
  \phi'\sigma_S\sigma_{L'}-\phi\sigma_S\sigma_L
  >e_S(e_L-e_{L'}).
  \label{eq:suppl_ceiling_change}
\end{equation}
\end{proposition}

\begin{proof}
A selector is wrong whenever both answers are wrong, so its accuracy is at most $1-\mathbb E[E_SE_L]$.
By the definition of covariance,
\begin{align*}
  \mathbb E[E_SE_L]&=e_Se_L+\operatorname{Cov}(E_S,E_L)\\
  &=e_Se_L+\phi\sigma_S\sigma_L.
\end{align*}
This proves the first claim.
The new bound is lower exactly when
$e_Se_{L'}+\phi'\sigma_S\sigma_{L'}>e_Se_L+\phi\sigma_S\sigma_L$.
Rearranging yields the second claim.
\end{proof}

The bound is reached by an oracle that selects a correct answer whenever one exists, so it is the oracle accuracy of the pair, before any fusion rule is chosen.
The proposition serves as a criterion for the view assignment.
Both sides of \cref{eq:suppl_ceiling_change} are measured from the answers of the two models. The right side is the accuracy that the new view adds to the second model, scaled by the error rate $e_S$ of the first, and the left side is the error covariance that it adds to the pair.
The diversity of ensemble members has been studied in relation to ensemble accuracy for classifiers that differ by training~\cite{kuncheva2003measures}.
Here both models are fixed, and the view assignment alone sets the covariance.
A more accurate \overseer lowers the oracle accuracy when its errors move onto the questions that the \scout already gets wrong.
Figure 6 of the main paper shows this case. When the \overseer also answers from the glimpse, it is correct on 204 more questions, while 26 fewer questions have at least one correct answer and the fused answer is correct on 33 fewer.
Over these 6,294 questions, the shared view lowers the error rate of the \overseer from 18.7\% to 15.5\%, raises the error correlation of the two models from 0.43 to 0.54 (95\% bootstrap interval of the rise over images $[0.06, 0.15]$), and raises the rate of joint errors by 0.41 points.
The covariance term thus grows by more than the product term shrinks, which is the condition of the proposition.

For the presence rule, define $U=\max_k\ell_k-\log\max(1,K/2)$.
Let the two component answers be yes exactly when $U>0$ and when $\ell_L(I)>0$, respectively.
The fused answer is yes exactly when $U+\ell_L(I)>0$.
When the component answers agree, the sum returns that answer, and when they disagree, the fused answer equals one of them.
The fused yes/no answer is therefore a selector, and the proposition applies with $E_S$ the error of the answer given by $U$.
For multiple-choice questions a product of experts can return an option that no view ranks first.
This occurs on 21 of the 6,184 escalated questions with DeepEyes-7B and on 28 with Thyme-RL, so the fused answer is the top answer of one of its views on more than 99.5\% of the escalated questions, and the proposition guides the view assignment.

\subsection{The Presence Count Correction}
\label{sec:suppl_presence}

Let $K$ be the number of available \scout views: the whole image, the drafted or recovered glimpse view when present, and the corrective glimpse view when present.
The presence statistic is
\begin{equation}
  U=\max_{1\leq j\leq K}\ell_j-\log\max(1,K/2).
  \label{eq:suppl_presence_stat}
\end{equation}
The maximum lets one view support presence without requiring that every view support it.
The following result gives the role of the count correction.

\begin{proposition}[False positives of the union]
\label{prop:suppl_presence_bound}
For each $k$ with $\Pr(Y=\text{no},K=k)>0$, assume that every available view satisfies
\begin{equation}
  \mathbb E[\exp(\ell_j)\mid Y=\text{no},K=k]\leq1,
  \quad j=1,\dots,k.
  \label{eq:suppl_null_moment}
\end{equation}
Then for any fixed real $\lambda$,
\begin{equation}
  \Pr(U>\lambda\mid Y=\text{no},K=k)
  \leq \min\{1,\min(k,2)e^{-\lambda}\}.
  \label{eq:suppl_presence_tail}
\end{equation}
The bound needs no independence between views.
\end{proposition}

\begin{proof}
Set $h=\log\max(1,k/2)$.
By Markov's inequality and the assumed moment, each $\ell_j$ exceeds $\lambda+h$ with probability at most $e^{-\lambda-h}$.
A union bound over the $k$ views gives $ke^{-\lambda-h}$.
Since $k/\max(1,k/2)=\min(k,2)$ and a probability is at most one, the claim follows.
\end{proof}

With the correction, the false-positive bound of three views equals that of two views, the setting of the two-view union.
The moment condition holds with equality when $\exp(\ell_j)$ is a likelihood ratio.
When $\ell_j$ is a posterior log-odds on the stratum $K=k$ with prior $\omega_k=\Pr(Y=\text{yes}\mid K=k)$, Bayes' rule writes $\exp(\ell_j)$ as the likelihood ratio times $\omega_k/(1-\omega_k)$.
The moment then equals the prior odds $\omega_k/(1-\omega_k)$, which is 1 on a balanced stratum.
The final answer adds $\ell_L(I)$ to $U$, so the \overseer can also outvote a false positive.
The effect of the correction is reported in \cref{tab:suppl_ablation_details}.

\subsection{Normalized Pooling and Score Order}
\label{sec:suppl_score_analysis}

With the same positive distributions $p_i$, define
\begin{equation}
  \bar p(a)=\frac{\exp\bigl(n^{-1}\sum_i\log p_i(a)\bigr)}{Z},
  \label{eq:suppl_normalized_pool}
\end{equation}
where $Z$ normalizes the distribution.

\begin{proposition}[Normalized pool]
\label{prop:suppl_normalized_pool}
For strictly positive $p_i$, the distribution in \cref{eq:suppl_normalized_pool} uniquely minimizes
\begin{equation}
  \frac{1}{n}\sum_{i=1}^{n}\operatorname{KL}(u\Vert p_i)
  \label{eq:suppl_pool_objective}
\end{equation}
over distributions $u$ on $\mathcal A$.
It has the same maximizers as the product used for multiple-choice answers, and replicating every member of $\mathcal V$ equally leaves it unchanged.
\end{proposition}

\begin{proof}
Expanding the objective gives
\begin{align*}
  \frac{1}{n}\sum_i\operatorname{KL}(u\Vert p_i)
  &=\sum_a u(a)\log u(a)\\
  &\quad-\sum_a u(a)\frac{1}{n}\sum_i\log p_i(a)\\
  &=\operatorname{KL}(u\Vert\bar p)-\log Z.
\end{align*}
The first term is nonnegative and vanishes only at $u=\bar p$.
Multiplying all log scores by the positive constant $1/n$ preserves their maximizers.
Equal replication leaves their average unchanged.
\end{proof}

The pool therefore keeps the scale of one view, however many views are fused.
For binary questions its log-odds is the mean log-odds of the pairs in $\mathcal V$.
The whole-image \scout distribution enters the pool when no drafted or recovered closer view takes its place.
The presence decision instead takes the maximum over all available \scout views, so the score $\bar p(a^\star)$ of a yes/no answer is the pooled probability of the answer that the union rule returned.

The trace score is
\begin{equation}
  r=\begin{cases}
  1+\min(c_0,c_1) & \text{certified exit},\\
  p_S(a_0\mid I) & \text{single-view exit},\\
  \bar p(a^\star) & \text{escalated answer}.
  \end{cases}
  \label{eq:suppl_score}
\end{equation}
A single-view exit has one pair in its trace, the \scout on the whole image, so its score is the pool of this pair, $p_S(a_0\mid I)$.
For $0<\tau\leq1$, the score of a certified exit exceeds one, while all other scores are at most one.
The certified exits therefore rank first, and single-view and escalated answers are ordered by their scores below them.
Every ranking metric depends only on this order.

\begin{table*}[t]
  \centering
  \caption{Paired tests of the accuracy gains in Table 1 of the main paper, with the image as the unit. Gain: mean accuracy over the six splits, in points. Better / Worse: images on which the method answers more / fewer questions correctly than the compared method. $p$: exact McNemar test on these images.}
  \label{tab:suppl_significance}
  \small
  \setlength{\tabcolsep}{7pt}
  \begin{tabular}{ll l c cc c}
  \toprule
  Expert & Method & Compared with & Gain & Better & Worse & $p$ \\
  \midrule
  \rowcolor{oursrow} DeepEyes-7B & \textbf{\method} & DeepEyes-7B agent & +4.9 & 321 & 94 & $4.2 \times 10^{-30}$ \\
  \rowcolor{oursrow}  & \textbf{\method} & SpecEyes & +3.1 & 157 & 64 & $3.4 \times 10^{-10}$ \\
   & \textbf{\methodfast} & DeepEyes-7B agent & +4.2 & 323 & 97 & $1.8 \times 10^{-29}$ \\
   & \textbf{\methodfast} & SpecEyes & +2.5 & 150 & 74 & $4.2 \times 10^{-7}$ \\
  \midrule
  \rowcolor{oursrow} Thyme-RL & \textbf{\method} & Thyme-RL agent & +3.7 & 312 & 115 & $4.3 \times 10^{-22}$ \\
  \rowcolor{oursrow}  & \textbf{\method} & SpecEyes & +1.8 & 170 & 99 & $1.8 \times 10^{-5}$ \\
   & \textbf{\methodfast} & Thyme-RL agent & +3.0 & 304 & 121 & $2.9 \times 10^{-19}$ \\
   & \textbf{\methodfast} & SpecEyes & +1.2 & 162 & 101 & $2.0 \times 10^{-4}$ \\
  \bottomrule
  \end{tabular}
\end{table*}

\begin{table*}[t]
  \begin{minipage}[t]{0.485\linewidth}
  \centering
  \caption{Accuracy (\%) of the Table 1 runs on POPE with the corrected labels of RePOPE.}
  \label{tab:suppl_repope_acc}
  \small
  \setlength{\tabcolsep}{4.5pt}
  \begin{tabular}{l ccc ccc}
  \toprule
   & \multicolumn{3}{c}{DeepEyes-7B} & \multicolumn{3}{c}{Thyme-RL} \\
  \cmidrule(lr){2-4} \cmidrule(l){5-7}
  Method & Rand. & Pop. & Adv. & Rand. & Pop. & Adv. \\
  \midrule
  Agent & 93.1 & 87.6 & 84.2 & 93.5 & 89.4 & 85.4 \\
  SpecEyes & 96.3 & 93.7 & 92.1 & 96.0 & 93.6 & 91.6 \\
  \rowcolor{oursrow} \textbf{\methodfast} & \second{96.6} & \second{94.5} & \second{93.3} & \second{96.6} & \second{94.2} & \second{93.2} \\
  \rowcolor{oursrow} \textbf{\method} & \textbf{96.8} & \textbf{94.7} & \textbf{93.7} & \textbf{96.7} & \textbf{94.4} & \textbf{93.4} \\
  \bottomrule
  \end{tabular}
  \end{minipage}\hfill
  \begin{minipage}[t]{0.485\linewidth}
  \centering
  \caption{AUROC (\%) of reliability scores for detecting wrong answers of \methodfast on POPE, with the original and the corrected labels, at batch size 1.}
  \label{tab:suppl_repope_trust}
  \small
  \setlength{\tabcolsep}{4.5pt}
  \begin{tabular}{l cc cc}
  \toprule
  AUROC $\uparrow$ & \multicolumn{2}{c}{DeepEyes-7B} & \multicolumn{2}{c}{Thyme-RL} \\
  \cmidrule(lr){2-3} \cmidrule(l){4-5}
  Score & POPE & RePOPE & POPE & RePOPE \\
  \midrule
  SpecEyes separability & 44.4 & 55.5 & 48.2 & 57.1 \\
  \scout single-view conf. & 56.4 & 65.4 & 57.0 & 65.4 \\
  \rowcolor{oursrow} \textbf{Trace reliability (ours)} & \textbf{80.5} & \textbf{89.6} & \textbf{80.4} & \textbf{89.0} \\
  \midrule
  \rowcolor{grouprow} \multicolumn{5}{l}{\textit{Questions sent to the \overseer}} \\
  \rowcolor{oursrow} \textbf{Trace reliability (ours)} & 84.1 & 89.9 & 84.9 & 90.1 \\
  \bottomrule
  \end{tabular}
  \end{minipage}
\end{table*}

\subsection{Bounded Stages and Cost}
\label{sec:suppl_cost_analysis}

Let $A_1,A_2,A_3$ indicate escalation, the glimpse check, and the corrective glimpse.
Both $A_2$ and $A_3$ imply $A_1$, so the number of \overseer stages is $N_L=A_1+A_2+A_3\leq3A_1$.
For $\Pr(A_1)>0$,
\begin{equation}
  \mathbb E[N_L]=\Pr(A_1)\bigl(1+\Pr(A_2\mid A_1)+\Pr(A_3\mid A_1)\bigr).
\end{equation}
A check stage batches one prompt per target label.

For model time, let $T_{\mathrm{init}}$ cover the initial \scout calls and $T_{\mathrm{esc}}$ the escalation calls, including recovered and corrective views, with $T_{\mathrm{esc}}=0$ outside escalation.
Then $T=T_{\mathrm{init}}+A_1 T_{\mathrm{esc}}$, so
\begin{equation}
  \mathbb E[T]=\mathbb E[T_{\mathrm{init}}]
  +\Pr(A_1)\mathbb E[T_{\mathrm{esc}}\mid A_1].
  \label{eq:suppl_cost}
\end{equation}
The cost of a question is bounded by three \overseer stages and four \scout calls, even after a rejection, and never includes the agent loop.
\method searches 13.2\% of the questions with DeepEyes-7B and 13.6\% with Thyme-RL.
On a searched question, it adds at most seven \overseer stages, each batched across questions, and two \scout calls, and the agent answers in one forward pass instead of running its loop.
Every early exit removes the second term.
\Cref{sec:suppl_impl} separates the end-to-end timing of Table 1 of the main paper from the model time of the batch-1 ablations.

\subsection{Empirical Error Bound of Certified Exits}
\label{sec:suppl_zero_errors}

The certified exits in \cref{fig:suppl_risk_coverage} contain no wrong answer.
A certified exit is a draft that has passed an empirical test on new evidence, and its reliability is a bound on the error rate of the exit rule at a stated confidence level, as in selective classification with guaranteed risk~\cite{geifman2017selective}.
The bound holds over the population of images, and each single answer carries its trace score.
The unit of the bound is the image.
HR-Bench asks each of its 200 questions under four orders of the options at both resolutions, and the three POPE splits share 500 images, so questions on one image are not independent draws.
We count the 4K and 8K versions of a picture as one image, which gives 842 images for the 10,791 questions.
Suppose the exit rule is fixed and the $n_c\geq1$ images with a certified exit are independent draws from the accepted population, and let $\epsilon$ be the probability that such an image has a wrong certified exit.
The probability of no errors is then $(1-\epsilon)^{n_c}$.
For $0<\eta<1$, inversion gives the one-sided upper confidence bound at level $1-\eta$ after zero errors,
\begin{equation}
  \epsilon\leq1-\eta^{1/n_c},
  \label{eq:suppl_zero_error_bound}
\end{equation}
which is the exact binomial bound of Clopper and Pearson~\cite{clopper1934use} for zero observed errors.
For any larger $\epsilon$, the probability of the zero-error event is less than $\eta$.
The 681 certified exits, 96 on V*, 573 on HR-Bench, and 12 on POPE, lie on 182 distinct images, and no image has a wrong exit.
At $\eta=5\%$, the bound is 1.63\%.
The test is also stable under the order of the options. On HR-Bench, the certified exits cover 166 pairs of a question and a resolution, and 122 of them exit under all four orders, each time with the correct answer.
At the released batch sizes of Table 1 of the main paper, 680 questions end at a certified exit with either expert, and all 680 answers are correct.
The exit rule has no parameter of its own. It compares two answers of the \scout, and its threshold $\tau$ is the released one of SpecEyes (\cref{sec:suppl_impl}).

\textbf{Cases of the trace score.}
The bound covers the certified exits, and the score in \cref{eq:suppl_score} places every other answer below them.
The accuracy falls across the three cases of the score: 100.0\% on the 681 certified exits, 93.9\% on the 3,926 single-view exits, and 84.5\% with DeepEyes-7B and 83.5\% with Thyme-RL on the escalated answers (\cref{tab:suppl_routes}).
The three cases hold 6.3\%, 36.4\%, and 57.3\% of the questions.
POPE has 9,000 of the 10,791 questions and an absent object has no region to zoom into, and the certified exits end 50.3\% of the V* questions and 35.8\% of the HR-Bench questions.
A single-view exit is a draft for which the \scout proposes no region, so the test has no second view to use.
On POPE, 3,491 of the 3,899 single-view exits answer that the object is absent.
By \cref{prop:suppl_exit_subset}, the wrong single-view exits are among the wrong answers of the confidence rule, which returns 700 wrong answers with no \overseer call on the same questions, against 240 for \methodfast (Table 2 of the main paper).

\textbf{Single-view exits and the agent loop.}
The single-view exits are also the better decision on their questions.
On the 3,926 questions that end at a single-view exit, the exits are 93.9\% correct, and the agent loop is 89.7\% correct with DeepEyes-7B and 89.9\% with Thyme-RL.
Of the 240 wrong exits, the agent loop is also wrong on 172 and 181, and of the 3,686 correct exits, it is wrong on 231 and 214.
Sending these questions to the agent would therefore turn more answers wrong than it corrects, and Table 2 of the main paper shows the same for the \overseer, as removing the early exit changes no column.

\subsection{Significance of the Accuracy Gains}
\label{sec:suppl_significance}

\Cref{tab:suppl_significance} tests the accuracy gains of Table 1 of the main paper.
All methods answer the same 10,791 questions, so each comparison is paired, and its unit is the image, as in \cref{sec:suppl_zero_errors}.
An image counts as better when the method answers more of its questions correctly than the compared method and as worse when it answers fewer, and the exact McNemar test~\cite{mcnemar1947note} is the two-sided binomial test on these two counts.
With several questions on one image, this is the sign test on images, and with one question per image, as on V*, it is the standard McNemar test on questions.
The gain is the difference in mean accuracy over the six splits, as in the last block of Table 1.
With both experts, \method and \methodfast are significantly more accurate than the agent and than SpecEyes.
The $p$-values against the agents are below $10^{-18}$, those of \method against SpecEyes are below $10^{-4}$, and the largest of the eight is $2.0 \times 10^{-4}$.
All eight tests remain significant at the 5\% level under the Holm correction for multiple comparisons~\cite{holm1979simple}.
Averaged over the two experts, every gain has a 95\% percentile interval above zero from 10,000 bootstrap resamples~\cite{efron1979bootstrap} of the images of each benchmark. The intervals are $[2.9, 5.7]$ and $[1.2, 3.8]$ points for \method over the agent and over SpecEyes, and $[2.2, 5.1]$ and $[0.5, 3.2]$ points for \methodfast.
V* has 191 questions, the fewest of the benchmarks, each on its own image, and both experts answer the same questions.
Resampling these questions, the gain of \method over the agent on V*, averaged over the two experts, is 4.7 points with a 95\% interval of $[0.5, 8.9]$.
The V* gain of \methodfast over the agent also holds with a 4B \scout (\cref{tab:suppl_scout_size}).

\subsection{POPE with Corrected Labels}
\label{sec:suppl_repope}

POPE takes its labels from the COCO annotations.
RePOPE~\cite{neuhaus2025repope} re-annotates the POPE images, corrects the wrong labels, and removes the ambiguous questions, which keeps 2,774, 2,727, and 2,684 questions of the random, popular, and adversarial splits.
We grade the same answers against these labels.
\Cref{tab:suppl_repope_acc} gives the accuracy of the Table 1 runs.
Both versions remain more accurate than the agent and SpecEyes on every split with both experts, and \method is the most accurate method on every split.
On the adversarial split, the gain of \method over SpecEyes grows from 1.3 points with the original labels to 1.6 and 1.8 points with the two experts.
\Cref{tab:suppl_repope_trust} gives the AUROC of the reliability scores.
With the corrected labels, the AUROC of the trace score rises from 80.5\% to 89.6\% with DeepEyes-7B and from 80.4\% to 89.0\% with Thyme-RL, and from 84.1\% and 84.9\% to 89.9\% and 90.1\% on the questions sent to the \overseer.
It stays the highest of the three scores, and its area under the risk-coverage curve falls from 3.11\% to 0.78\% and from 3.32\% to 0.85\%.
Of the 234 wrong single-view exits on POPE, RePOPE corrects the label of 137, so that the exit answer is correct, and removes 41 as ambiguous.
With the corrected labels, the single-view exits on POPE are 98.0\% correct, and the 12 certified exits remain correct.

\end{document}